\documentclass{article}

\PassOptionsToPackage{sort&compress, numbers}{natbib}

\usepackage[final]{neurips_2024}
\usepackage{graphicx}
\usepackage{wrapfig}
\usepackage{tabularx}
\usepackage{multicol}
\usepackage{subcaption}
\usepackage{tabularx}
\usepackage[only,llbracket,rrbracket]{stmaryrd}
\usepackage{listings}

\usepackage{amsmath,amsfonts,bm}

\def\eqref#1{equation~\ref{#1}}

\def\floor#1{\lfloor #1 \rfloor}
\def\1{\bm{1}}

\def\rmD{{\mathbf{D}}}
\def\rmE{{\mathbf{E}}}

\def\rmH{{\mathbf{H}}}

\def\rmU{{\mathbf{U}}}

\def\vh{{\bm{h}}}

\def\vx{{\bm{x}}}

\def\vz{{\bm{z}}}

\DeclareMathAlphabet{\mathsfit}{\encodingdefault}{\sfdefault}{m}{sl}
\SetMathAlphabet{\mathsfit}{bold}{\encodingdefault}{\sfdefault}{bx}{n}

\def\gA{{\mathcal{A}}}

\def\sB{{\mathbb{B}}}

\def\sH{{\mathbb{H}}}
\def\sI{{\mathbb{I}}}

\def\sL{{\mathbb{L}}}

\def\sO{{\mathbb{O}}}

\def\sR{{\mathbb{R}}}

\def\sZ{{\mathbb{Z}}}

\newcommand{\ctexttt}[1]{\texttt{\textcolor{NavyBlue}{#1}}}

\usepackage[utf8]{inputenc} 
\usepackage[T1]{fontenc}    
\usepackage[hidelinks]{hyperref}       
\usepackage{url}            
\usepackage{booktabs}       
\usepackage{amsfonts}       
\usepackage{amssymb}
\usepackage{nicefrac}       
\usepackage{microtype}      
\usepackage[dvipsnames]{xcolor}         

\definecolor{CambridgeRed}{HTML}{D6083B}
\definecolor{CambridgeGreen}{HTML}{55A51C}
\definecolor{CambridgeBlue}{HTML}{0072CF}

\title{Deep Equilibrium Algorithmic Reasoning}

\author{%
Dobrik Georgiev\\
University of Cambridge\\
{\tt\small dgg30@cam.ac.uk}\\
\And
JJ Wilson\\
Independent Researcher\\
{\tt\small josephjwilson74@gmail.com}\\
\AND
\hspace{-30pt}
Davide Buffelli\\
\hspace{-30pt}
MediaTek Research \\
\hspace{-30pt}
{\tt\small davide.buffelli@mtkresearch.com}\\
\And
\hspace{-26pt}
Pietro Liò \\
\hspace{-26pt}
University of Cambridge \\
\hspace{-26pt}
{\tt\small pl219@cam.ac.uk} \\
}

\begin{document}

\maketitle

\begin{abstract}
Neural Algorithmic Reasoning (NAR) research has demonstrated that graph neural networks (GNNs) could learn to execute classical algorithms. However, most previous approaches have always used a recurrent architecture, where each iteration of the GNN matches an iteration of the algorithm. In this paper we study neurally solving algorithms from a different perspective: since the algorithm's solution is often an equilibrium, it is possible to find the solution directly by solving an equilibrium equation. Our approach requires no information on the ground-truth number of steps of the algorithm, both during train and test time. Furthermore, the proposed method improves the performance of GNNs on executing algorithms and is a step towards speeding up existing NAR models. Our empirical evidence, leveraging algorithms from the CLRS-30 benchmark, validates that one can train a network to solve algorithmic problems by directly finding the equilibrium. We discuss the practical implementation of such models and propose regularisations to improve the performance of these equilibrium reasoners.
\end{abstract}
\footnotetext[0]{Source code available here: \url{https://github.com/HekpoMaH/DEAR}}

\section{Introduction}

Algorithms, while straightforward in theory, become challenging to deploy in
real-world scenarios. They operate in abstract domains with very specific
conditions and types of inputs, which are represented with scalars.
The main hurdle is the need to ``collapse'' reality into a scalar every time an
algorithm is used, something usually done based on intuition rather than
principled science \citep{harris1955fundamentals}. Neural Algorithmic Reasoning
(NAR;\@ \citealp{velickovic2021neural}) has been proposed to address this issue
by utilising specialised neural network architectures to break this scalar
bottleneck by executing algorithms in higher-dimensional space made out of
arrays of numbers (vectors). The task of mapping reality into this vectorial
space is delegated to automated gradient-based optimisation techniques rather
than relying on human operators.

While NAR does not provide the correctness guarantees of its classical
counterparts, robust performance can be achieved through \textit{alignment}
\citep{xu2020what} -- submodules of the model architecture correspond to
easy-to-learn subparts of the algorithm (or class of). Graph Neural Networks
(GNNs) have emerged as the most convenient architecture to execute all types of
algorithms \citep{ibarz2022generalist} and GNNs that align better to the target
algorithm achieve better generalisation.  This alignment game
\citep{velickovic2023NARGradient} has led to a sequence of exciting research --
from aligning the architecture with iterative algorithms
\citep{tang2020towards} to proving that ``graph neural networks are dynamic
programmers'' \citep{dudzik2022graph}, especially if their message-passing
function \citep{gilmer2017neural} takes into account 3-node interactions.

The aforementioned papers focus on aligning the computation of the GNN
with an algorithm or a specific algorithm class (e.g.\ dynamic programming),
but ignore the properties \emph{at the time of algorithm termination}. For the
algorithms in the CLRS-30 algorithmic reasoning benchmark
\citep{velivckovic2022clrs} once the optimal solution is found, further
algorithm iterations will not change it. For example, in dynamic programming
shortest-paths algorithms making additional iterations would not alter the
optimality of the shortest paths' distances found. Such a state is called an
\emph{equilibrium} -- additional applications of a function (an algorithm's
iteration) to the state leave it unchanged.


In this paper:
\begin{enumerate}
    \item We explore the connection between execution of algorithms and
        equilibrium finding through the use of Denotational semantics and
        Domain theory. (\autoref{subsec:dsadt}) 
    \item Inspired by the above, we implement a class of deep equilibrium
        algorithmic reasoners (DEARs) that learn algorithms by identifying the
        equilibrium point of the GNN equation and propose improvements to them.
        (\autoref{sec:DER})
    \item Our results suggest that the above reasoners can be
        \emph{successfully} trained. Not only does equilibrium algorithmic
        reasoning achieve better overall performance with less expressive (and
        expensive) GNNs, but is also competitive to the more expressive (and
        expensive) NAR models. All this is done without providing any
        information on the number of algorithmic steps -- neither at train nor
        at test time.  (\autoref{sec:experimental})
    \item DEARs also drastically improve the inference speed -- an achievement
        made possible by the use of optimised root-finding algorithms and by
        decoupling the neural model from the \emph{sequential} implementation
        of algorithms in standard benchmark datasets. (\autoref{sec:experimental})
\end{enumerate}

\paragraph{Related work}
The main proposed application of NAR is settings where one wants to apply an
algorithm, but it is impossible to represent reality with a single scalar, hence
an ``executor'' operating in vector space and faithful to the algorithm is
required \citep{NARBlueprint, velickovic2023NARGradient}. As NAR models are
neural clones of algorithms, they need to provide correct output even for
previously unobserved input sizes. Achieving robust out-of-distribution (OOD)
generalisation is tricky. To this end a plethora of works have dedicated their
attention to it -- \citep{tang2020towards, dudzik2022graph,
bevilacqua2023neural, dudzik2024async} to name a few. Except for one concurrent work
(a blog post;\citep{xhonneux2024deepequilibriummodels}), those works focus on
improving the GNN step and many completely ignore the termination of algorithms
or any properties of the last state, such as equilibrium. This work, similarly
to \citet{xhonneux2024deepequilibriummodels}, studies neurally finding
solutions to algorithms by relying on the equilibrium property. We, however,
attempt to give the precise assumptions required for this approach to work.
Further, we implement a more robust model than theirs, which achieves
comparable or better performance to baselines. Finally, we propose
modifications to improve equilibrium NARs.


\section{Background}\label{sec:background}

\paragraph{Algorithmic Reasoning}
Let $A: \sI_A \to \sO_A$ be an algorithm, acting
on some input $\vx \in \sI_A$, producing an output $A(\vx) \in \sO_A$ and let
$\sI_A$/$\sO_A$ be the set of possible inputs/outputs $A$ can read/return. In
algorithmic reasoning, we aim to learn a function $\gA: \sI_A \to \sO_A$, such
that $\gA \approx A$.  Importantly, we will not be learning simply an
input-output mapping, but we will aim to align to the algorithm $A$'s
trajectory. The alignment is often achieved through direct
supervision\footnote{Recent research \citep{bevilacqua2023neural,
rodionov2023neural} has shown that alternative, causality-inspired, ways of
alignment also exist.} on the intermediate states of the algorithm. To capture
the execution of $A$ on an input $x$ we can represent it as
\begin{subequations}
  \begin{tabularx}{1.0\textwidth}{p{3.65cm}Xp{4.4cm}}
    \begin{align}
    \bar{\vh}_0 = \text{\textsc{Preproc}}(\vx)
    \end{align}
    &
    \begin{align}
    \bar{\vh}_\tau = \underbrace{A_t(\dots A_t}_\text{$\tau$ times}(\bar{\vh}_0)\dots)\label{eq:algorollout}
    \end{align}
    &
    \begin{align}
    A(\vx) = \textsc{Postproc}(\bar{\vh}_\tau) 
    \end{align}
  \end{tabularx}
\end{subequations}
where \textsc{Preproc} and \textsc{Postproc} are some simple pre- and
post-processing (e.g.initialising auxiliary variables or returning the correct
variable), $\bar{\vh}_\tau\in \sH_A$ is $A$'s internal (\textbf{h}idden) state,
$A_t$ is a subroutine (or a set of) that is executed at each step and the
number of steps depends on a boolean property being satisfied (e.g. all nodes
visited). It is therefore no surprise that the encode-process-decode
architecture \citep{hamrick2018relational} is the de-facto choice when it comes
to NAR.  Thus, the architecture can be neatly represented as a composition of
three learnable components: $\gA= g_\gA \circ P \circ f_\gA$, where $g_\gA:
\sI_A \to \sR^d$ and $f_\gA: \sR^d \to \sO_A$ are the encoder and decoder
function respectively (usually linear projections) and $P: \sR^d \to \sR^d$ is
a processor that mimics the rollout (\autoref{eq:algorollout}) of $A$. The
processor often uses a message-passing GNN at its core.

\paragraph{CLRS-30} The \emph{CLRS-30} benchmark \citep{velivckovic2022clrs}
includes 30 iconic algorithms from the \textit{Introduction to Algorithms}
textbook \citep{CLRS}. Each data instance for an algorithm $A$ is a graph
annotated with features from different algorithm stages (\emph{input},
\emph{output}, and \emph{hint}), each associated with a location (\emph{node},
\emph{edge}, and \emph{graph}). Hints contain time series data representing the
algorithm rollout and include a temporal dimension often used to infer the
number of steps $\tau$. Features in CLRS-30 have various datatypes with
associated losses for training. The test split, designed for assessing
out-of-distribution (OOD) generalisation, comprises graphs four times larger
than the training size.

\paragraph{Deep Equilibrium Models}
Deep equilibrium models \citep[DEQs][]{bai2019deep} are a class of
implicit neural networks \citep{ghaoui2019implicit}. The functions modelled
with DEQs are of the form:
\begin{align}
    \vz^*=f_\theta(\vz^*, \vx)\label{eq:fixedpoint}
\end{align}
where $\vx$ is input, $f_\theta$ is a function parametrised by $\theta$ (e.g.
a neural network) and $\vz^*$ is the output. $\vz^*$ is an equilibrium point to
the eventual output value of an infinite depth network where each layer's
weights are shared, i.e.  $f_\theta^{[i]}=f_\theta$. By re-expressing
(\ref{eq:fixedpoint}) as $g_\theta=f_\theta(\vz^*, \vx)-\vz^*$ DEQs allow us to
find the fixed point $\vz^*$ via any black-box root-finding method
\citep[e.g.][]{broyden1965aclass, anderson1965iterative}, without the actual
need to unroll the equation until convergence, allowing us to reduce steps.
For backpropagation the gradient $\partial \mathcal{L}/\partial \theta$ could
be calculated using the Implicit Function Theorem (cf. \citealp{bai2019deep})
and no intermediate state has to be stored, giving us a constant memory cost of
gradient computation regardless of the number of iterations until convergence.



\paragraph{Expander graphs}
MPNNs operate by exchanging information between adjacent nodes
\citep{gilmer2017neural}. It has been identified that the message passing
process can be hindered by a phenomenon known as \emph{oversquashing}
\citep{alon2020bottleneck}, which occurs when a large volume of messages are
compressed into fixed-sized vectors. The importance of overcoming the negative
implication posed by this phenomenon is crucial for the overall expressivity of
GNNs \citep{giovanni2023oversquashing}, particularly in the context of
long-range node interactions.


To this end, several spectral methods have been proposed to mitigate
oversquashing by increasing the Cheeger constant \citep{arnaiz2022diffwire,
banerjee2022oversquashing, karhadkar2022fosr}. A higher Cheeger constant
provides a measurement that a graph is globally lacking bottlenecks. The novel
approaches include graph rewiring techniques \citep{topping2021understanding,
barbero2024localityaware}, as well as significant independent bodies of research
recognising the efficacy of expander graphs \citep{deac2022expander,
shirzad2023exphormer, banerjee2022oversquashing}, due to their desirable
properties.



Expander graphs are proven to be highly connected sparse graphs ($|E| = \mathcal{O}(|V|)$) with a low diameter \citep{mohar1991eigenvalues}, thus offering advantageous properties for information propagation. Consequently, this facilitates messages to be passed between any pair of nodes in a short number of hops, and as a result,
alleviating oversquashing. 
Formally, a graph $G = (V, E)$ is defined as an
expander if it satisfies certain expansion properties. One common definition
involves the aforementioned Cheeger constant.  In the work of
\citet{deac2022expander}, a high Cheeger constant is equivalent to a graph
being bottleneck free \citep[Definition 3]{deac2022expander}, and that an
expander has a high Cheeger constant \citep[Theorem 5]{deac2022expander}.





There are various methods for constructing expander graphs. We opt for the
\emph{deterministic} algebraic approach as in \citet{deac2022expander},
utilising Cayley graphs. Specifically, we leverage Definition 8 and Theorem
9 of \citep{deac2022expander} to construct the Cayley graph for the
\emph{special linear group} $\mathrm{SL}(2,\sZ_n)$ and its generating set $S_n$
-- see p.5 of \citet{deac2022expander} for details.  
Note, that the order of 
a Cayley graph for $\sZ_n$ is $\left|V\right|=\mathcal{O}(n^3)$.
Hence, for many input graphs, a Cayley graph of the same size may not exist.

\section{Denotational semantics: The denotation of a while loop statement}\label{subsec:dsadt}

This aim of this section is to draw the parallel between finding equilibriums
and executing an algorithm, in order to answer if and when an equilibrium NAR
model can be successful. The following paragraphs introduce the mathematical
tools for formalising \emph{fixed points} -- denotational semantics
\citep{scott1971toward} and domain theory \citep{scott1982domains}. 

\paragraph{Denotational semantics} To every programming language
expression\footnote{Note the abuse of notation. For this subsection, we will forget
we use $P$ for the neural processor.} $P$ denotational semantics provides
an interpretation $\llbracket P \rrbracket$, which is a mathematical object
(often a function), representing the behaviour of the expression to
different inputs. These \emph{denotations} must be: 1) abstract, i.e.
independent of language and hardware, hence functions are natural choice; 2)
compositional, i.e. $\llbracket P \rrbracket$ can only be defined in terms of
the \emph{denotations} of $P$'s subexpressions, but not $P$ itself; 3) model
the computation $P$ performs.  As an example, we will use the lightweight
imperative programming language \textbf{IMP}\footnote{Due to space constraints,
we omit the formal language definition (\citet{winskel1993formal}, p. 11-13),
and we use a condensed version \citep{fioredenotational} of the denotational
syntax, given in Chapter 5 of \citet{winskel1993formal}.}. It consists of
\textit{numbers}, \textit{locations}, \textit{arithmetic expressions},
\textit{boolean expressions}  and \textit{commands}. Examples of \textbf{IMP}
are given in \autoref{app:ebigo2} -- we will use \textcolor{NavyBlue}{blue} for
\textbf{IMP} and encourage the reader to check how we use colour in the
appendix. Albeit small, the language is Turing-complete and all algorithms we
experiment with can be defined in it. 

Denote the set of variable locations with $\sL$ -- those
are all variables/array elements we can ever define. A good analogy to think of
is the addresses in the language \texttt{C}. The notation we can use to
represent a program state is $[X\mapsto 1, B\mapsto -48, \dots]$ and means that
the value of $X$ is 1, the value of $B$ is $-48$ and so on. In other words,
program states map locations to integers, s.t.\ a location can be mapped only
once. Hence states are functions and the set of all program states
$State$ is the set of functions mapping locations to integers: given $s \in
State$, $s(L)\in \sZ$ is the value at the location $L$ \emph{for the state
$s$}. The value for location $L$ in a different $s'\in State$, $s'(L)$, may or
may not differ.  The denotation of arithmetic / boolean expressions / commands
are the functions with domain $State$. These will be represented in the form of
\emph{lambda abstractions}, i.e. $\lambda x\in S. M$ rather than $f(x\in S)=M$,
where $S$ is a set and $M$ is a function body. The codomain of the denotation
depends on the type of expression: $\llbracket a \rrbracket : State \to \sZ$
for arithmetic expressions, $\llbracket b \rrbracket : State \to \sB$, for
boolean expressions and $\llbracket c \rrbracket : State \rightharpoonup State$
for commands. Since commands transform state, they are also called state
transformers. \textbf{\emph{Commands' denotations are partial functions, as
expressions like}} \ctexttt{while true do skip} \textbf{\emph{never terminate
and have no denotation.}}



For a large portion of the above language, it is trivial and intuitive to
define the denotations by structural recursion. For
example:
\begin{align*}
    &\llbracket \ctexttt{if}\ b\ \ctexttt{then}\ c_0\ \ctexttt{else}\ c_1 \rrbracket = \lambda s\in State.\
    \begin{cases}
        \llbracket c_0 \rrbracket(s) \quad \text{if }\llbracket b \rrbracket(s) \text{ is \ctexttt{true}}\\
        \llbracket c_1 \rrbracket(s) \quad \text{otherwise}
    \end{cases}\\
    \llbracket \ctexttt{(}c_0\ctexttt{;}c_1\ctexttt{)} \rrbracket &= \lambda s\in State.\ \llbracket c_1 \rrbracket(\llbracket c_0 \rrbracket(s)) \qquad\qquad\qquad \llbracket skip \rrbracket = \lambda s \in State.\ s
\end{align*}
The only denotation that cannot be expressed recursively, is that of the
\ctexttt{while} construct. Let $w = \ctexttt{while}\ b\ \ctexttt{do}\ c$. By
program equivalence, $w = \ctexttt{if}\ b\ \ctexttt{then}\ \ctexttt{(}c\ctexttt{;}w\ctexttt{)}\ \ctexttt{else
skip}$. Therefore
\begin{align*}
    \textcolor{CambridgeRed}{\llbracket w \rrbracket} = \llbracket \ctexttt{if}\ b\ \ctexttt{then}\ \ctexttt{(}c\ctexttt{;}w\ctexttt{)}\ \ctexttt{else skip} \rrbracket &= \lambda s\in State.\
    \begin{cases}
        \textcolor{CambridgeRed}{\llbracket w \rrbracket}\left(\llbracket c \rrbracket(s)\right) & \text{if }\llbracket b \rrbracket(s) \text{\ is \ctexttt{true}}\\
        s & \text{otherwise}
    \end{cases}
\end{align*}
but this is not a valid definition, since it reuses $\llbracket w \rrbracket$
(highlighted in \textcolor{CambridgeRed}{red} above). Denotational semantics solves this
problem, by defining a function $f_{b,c}: (State \rightharpoonup State) \to
(State \rightharpoonup State)$ which takes one state transformer and returns
another:
\begin{align}
f_{b,c} = \lambda \hat{w} \in (State \rightharpoonup State). \lambda s \in State.\
    \begin{cases}
        \hat{w}\left(c(s)\right) & \text{if }b(s)\\
        s & \text{otherwise}
    \end{cases}\label{eq:fbc}
\end{align}
$\hat{w}$ is now a function variable. The denotation of $\llbracket
w \rrbracket$ is the fixed point of $f_{\llbracket b \rrbracket, \llbracket
c \rrbracket}$, i.e.\ $\llbracket w \rrbracket = f_{\llbracket b \rrbracket,
\llbracket c \rrbracket} (\llbracket w \rrbracket)$.  In order
to find the denotation, we need to solve the fixed point.  To aid the reader
a full worked example of computing the denotation for a \ctexttt{while} loop
construct is given in \autoref{app:ebigo}.

\paragraph{Domain theory}  \citet{scott1982domains} provides a framework with
which we can both find and also characterise solutions for fixed point
equations.\footnote{For detailed definitions and proofs, please refer to \S5.4
of \citet{winskel1993formal}.} Define $D$ as the domain of state transformers
$(State \rightharpoonup State)$. A partial order\footnote{It is reflexive,
transitive and anti-symmetric.} $\sqsubseteq$ on $D$ is defined as follows:
$w \sqsubseteq w'$ iff $\forall s \in State$ if $w(s)$ is defined then
$w(s)=w'(s)$. In other words $w'$ keeps old mappings and only defines new
ones. The totally undefined partial function $\bot$ is the least element in
$D$. This function contains no location to value mappings. A chain is
a sequence of elements of $D$, s.t.  $d_0 \sqsubseteq d_1 \sqsubseteq d_2
\sqsubseteq \dots$ . The supremum of the chain, called \emph{least upper
bound (lub)}, is denoted as $\bigsqcup_{n\geq0} d_n$. There could exist
different chains, but, by definition, all chains in a domain must have
a lub.

A function $f: D \to D$ is monotonic iff $\forall d,d'\in D.\ (d \sqsubseteq d'
\Rightarrow f(d) \sqsubseteq f(d'))$. In other words, if the second state
defined more mappings than the first and we apply one iteration step to both
states, the state resulting from the second will still define more mappings.
Monotonic functions for which $\bigsqcup_{n\geq0} f(d_n)
= f(\bigsqcup_{n\geq0}d_n)$ are also called continuous. In plain language, if
a function is continuous and we are provided with a chain, the lub of $f$
applied to every chain element is the same as $f$ applied to the lub of the
chain. An element $d''\in D$ is defined to be \emph{pre-fixed point } if
$f(d'')\sqsubseteq d''$ -- applying $f$ does not define anything new. The fixed
point $fix(f)$ of $f$ is the least pre-fixed point of $f$. By utilising
antisymmetry\footnote{$a \sqsubseteq b$ and $b \sqsubseteq a$ implies $a=b$} of
$\sqsubseteq$ and the two properties of $fix(f)$ (pre-fixed point and least) we
can obtain $f(fix(f))=fix(f)$. By Tarski's theorem \citep{tarski1955lattice},
any continuous $f: D \to D$ has a least pre-fixed point. This fixed point can
be found, by taking the lub of the chain of applications of f: $fix(f)
= \bigsqcup_{n\geq0}f^n(\bot)$. The helper function $f_{b,c}$ from
\autoref{eq:fbc} is continuous \citep[proof is given on p.120
of][]{gunter1992semantics}, therefore a direct result is that if the
$\ctexttt{while}\ b\ \ctexttt{do}\ c$ terminates then its denotation exists and
is \emph{the least} fixed point of sequence of iterations (compared to picking
any fixed point).

\paragraph{Denotational semantics and NAR}
The above detour into denotational semantics has affirmed the existence of
a connection between equilibrium models and algorithms (as conjectured by
\citet{xhonneux2024deepequilibriummodels}). Under the assumptions that:
\begin{itemize}
    \item the algorithms always terminate -- while not computable in the
        general case, this holds true for experiments, as we are dealing with
        offline algorithms with provable termination
    \item the algorithms we train on can be modelled as ``$\ctexttt{while}\ b\
        \ctexttt{do}\ c$'' constructs within \textbf{IMP}
\end{itemize}
the least fixed point exists and can be found by taking the first ``state'' of
the algorithm where future iterations on it have no effect.  In
\autoref{app:equilibriums} we have further annotated three algorithms from the
official code of the CLRS benchmark: BFS, Floyd-Warshall, strongly connected
components. Those annotations clearly show that algorithms can be rewritten in
\textbf{IMP} regardless of their implementation size. While BFS is clearly
a ``$\ctexttt{while}\ b\ \ctexttt{do}\ c$''-type of algorithm, annotating the
other two reveals that either the network may need more input features to decide
termination (Floyd-Warshall; \autoref{app:lst:fw}) or that the algorithm can be
composed of several while loops where each $c$ is another while loop on its own
(strongly connected components; \autoref{app:lst:scc}).  Fortunately, our
approach is not doomed to fail: a single DEQ layer can model any number of
``stacked'' DEQ layers \citep[chapter 4]{kolter2023deep}.


\section{Deep equilibrium algorithmic reasoning}\label{sec:DER}



\paragraph{Architecture} We implement our processors/encoders/decoders following
\citet{ibarz2022generalist}. The most notable difference\footnote{See
\autoref{app:differencesCLRS} for others not mentioned in the main text} to
their implementation is that ours uses a sparse graph representation. This
requires us to assume a fully connected graph on tasks where no graph structure
exists, in order to be able to give pointer predictions, and to reimplement the
strongly connected components algorithm so that the output pointers are always
in the edge set (this did not change the difficulty of the task).

The final node embeddings, from which the output is decoded, are the solution to
the equation:
\begin{align}
    \begin{split}
        \rmH^{(*)}=P_{\rmU\rmE}(\rmH^{(*)})
    \end{split}\label{eq:DER}
\end{align}
where $P_{\rmU\rmE}(\rmH^{(*)})=P(\rmH^{(*)}, \rmU, \rmE)$, $\rmU$/$\rmE$ are
the encoded node and edge feature matrices and $P$ is the processor function.
$\rmH^{(t)}$ are the stacked latent states of the nodes at timestep $t$ (with
$\rmH^{(0)}=\mathbf{0}$). The above \autoref{eq:DER} matches the signature of
\autoref{eq:fixedpoint}, and can be solved via root-finding (we employ
\texttt{torchdeq} \citep{torchdeq}; MIT License), as if it is $f_\theta$ of
a DEQ. Any model using this technique will be called \emph{deep equilibrium
algorithmic reasoner} (\textbf{DEAR}) in our experiments. The default
processor in the majority of our experiments is a PGN
\citep{velickovic2020pointer} with a gating mechanism as in
\citet{ibarz2022generalist}, but we note that DEARs can use any kind of
processor.


\paragraph{Finding the fixed point} The \texttt{torchdeq} library provides
several solvers. The most basic one is \emph{fixed-point iteration},
equivalent to repeating \autoref{eq:DER} until convergence.
However, in our very first experiments the solver needed more iterations than
the algorithm we train on. We therefore opted for the \emph{Anderson} solver
(implements Anderson acceleration \citep{anderson1965iterative}) 
and abandoned fixed-point iteration: 
\begin{align*}
    \hat{\rmH}^{(t+1)} = P_{\rmU\rmE}(\rmH^{(t)})\qquad
    \rmH^{(t+1)} = SolverStep\left(\left[ \rmH^{(0)} \dots \rmH^{(t)}\right], \hat{\rmH}^{(t+1)}\right)
\end{align*}
The criteria for pre-fixed point check \texttt{torchdeq} implements is
distance-based: for a state $\rmH^{(t)}$ to be considered a pre-fixed point,
the distance to the next state has to be under a pre-defined threshold
$\delta$. We kept the criteria but modified the library to always return the
least pre-fixed point (see \autoref{app:lfp}). This is in contrast to picking
the pre-fixed point with the least distance to next state (the default option
in \texttt{torchdeq}) and is a decision largely motivated from
\autoref{subsec:dsadt}. Due to a lack of a suitable definition for the domain
of NAR trajectories, we define $\forall t.\ \rmH^{(t)}\sqsubseteq\rmH^{(t+1)}$,
i.e.\ we pick the first state that passes the pre-fixed point check.


\paragraph{Globally propagating information} For problems defined on graphs it
is in theory possible that the number of solver iterations needed to find
equilibrium is less than the diameter of the graph. While, in practice, this is
unlikely to happen we hypothesise that improving long-range interactions could
improve the convergence of DEAR.  For this
reason, we adopt the implementation of Cayley Graph Propagation (CGP)
\citep{wilson2024cayley}. Contrasted to Expander Graph Propagation (EGP) \citep{deac2022expander}, which addresses the graph size misalignment (see \autoref{sec:background}) by truncating the Cayley graph, CGP keeps the extra nodes as virtual nodes. The CGP model upholds the aforementioned advantageous properties of an expander graph in a more grounded manner by preserving the complete structure.

In GNNs, the benefits of augmenting a graph with virtual nodes and providing
message-passing shortcuts have been observed to improve performance in various
tasks \citep{hu2020open, hu2021ogb, cai2023connection}; further supported by
the theoretical analysis \citep{hwang2022analysis}. Additionally, by retaining
the complete Cayley graph structure we improve the structure-aware
representations by varying the neighbourhood ranges
\citep{xu2018representation}.



\paragraph{No hint by default} We do not make any use of hints (supervising on
intermediate algorithm state). First, although it may seem counterintuitive, it
has been shown that a NAR model can successfully generalise, and even give
better results when trained to only predict the correct output
\citep{bevilacqua2023neural, rodionov2023neural}. Second, the fact that the
solver uses the GNN exactly once per call \emph{does not imply that one step of
the solver would correspond to one iteration of the algorithm}, bringing
uncertainty which DEAR states to match to which algorithm step. While we
propose an alignment scheme (see next paragraph), which has the potential to
integrate hints, we leave this for future work.

\begin{figure}
    \centering
    \includegraphics[width=\linewidth]{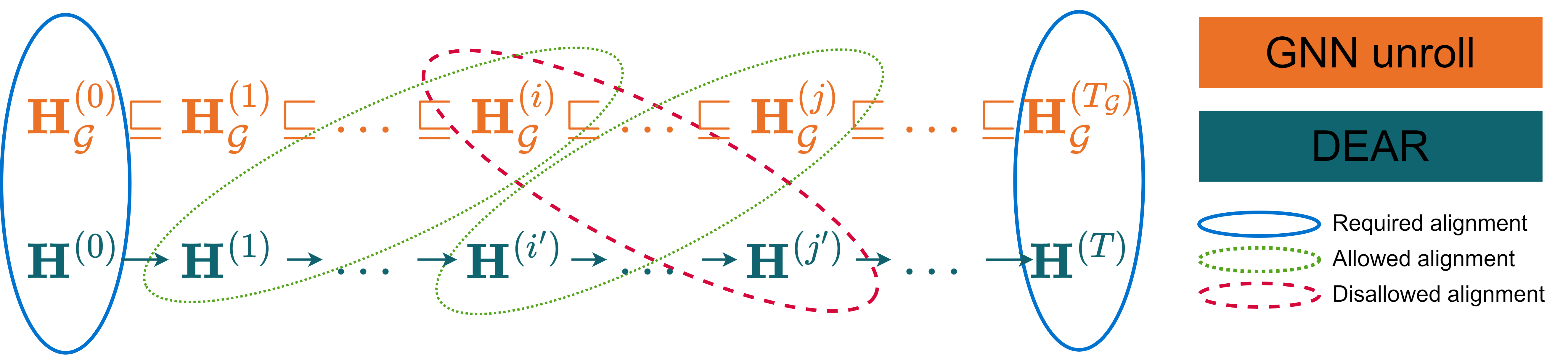}
    \caption{%
        Proposed alignment rule: every state in the DEAR trajectory should ``go
        forward''. Alignments to a GNN state that has already been ``passed''
        are disallowed. First and last states must always align. We
        intentionally use arrows instead of $\sqsubseteq$ for DEAR,
        as $\sqsubseteq$ may not hold for DEAR's trajectory.
    }\label{fig:align_example}
\end{figure}

\paragraph{Alignment} Our idea of alignment is visualised in
\autoref{fig:align_example}. We are given two trajectories of states, one
obtained from unrolling GNN iterations as in classical NAR and another obtained
from using DEAR. We would like to match DEAR to NAR, such that $\forall i\leq
j, i'\leq j'$ if we have committed to aligning DEAR state $\rmH^{(i')}$ to NAR
state $\rmH^{(j)}_\mathcal{G}$, we cannot align any $\rmH^{(j')}$ to
$\rmH^{(i)}_\mathcal{G}$ and from the same start we would like to reach the
same final state. In other words, \emph{skipping states is allowed, going back
in time is not}. This auxiliary supervision would also improve the monotonicity
of DEARs, encouraging faster convergence. Enforcing this alignment is done by
using an auxiliary loss. Choosing the $L_2$ norm as a distance metric, we use
a dynamic programming algorithm (\autoref{app:alignDP}) to compute the most
optimal alignment (normalised by the number of solver steps, in order not to
penalise longer trajectories) and supervise on that value.

Even with normalisation, the alignment sometimes had the effect of making the
optimisation stuck in local minima where the number of steps to hit equilibrium
was as low as 2 and the gradients were uninformative. We combated this in two
ways: 1) instead of using the default layer normalisation we switched to
\textsc{granola} \citep{eliasof2024granola}; 2) since $f(fix(f))=f$, we
performed a random number of additional iterations \citep{liu2024scalable} and
take the last state. The probability of doing $s$ extra steps is $0.5^s$.


\section{Evaluation}\label{sec:experimental}

\paragraph{Setup} For each algorithm we generated $10^5$/100/100-sized
training/validation/test datasets. Training sample sizes vary between 8 and 16
elements (uniformly randomly chosen) validation samples are of size 16. As is
standard in NAR literature, we measure the test performance
out-of-distribution, so our test samples are of size 64. For algorithms on
graphs we generate Erd\H{o}s–Rényi graphs \citep{erdos1960evolution} with edge
probabilities $p$ uniformly sampled from the interval $[0.1, 0.9]$, with
increments of $0.1$, which is the data distribution our baselines
\citep{ibarz2022generalist, bevilacqua2023neural} have used. We obtained the
ground truth execution trajectories and targets using the CLRS-30
implementation \citep{velivckovic2022clrs}.

In our experiments the models have a latent dimensionality of 128, the batch
size is 32, the learning rate is $3\times10^{-4}$ and we use the Adam optimizer
\citep{kingma2015adam}. We train our algorithmic reasoners for 100 epochs,
choosing the model with the lowest \emph{task} validation loss (discounting any
regularisation; focusing on performance only) for testing. Each task is
independently learned, minimising the output loss (losses depend on the
algorithm, cf. CLRS-30) plus any regularisation losses. Unless otherwise
specified, DEARs employ the Anderson root-finding method from the
\texttt{torchdeq} library and include Jacobian regularization
\citep{bai2021stabilizing}, the tolerance for fixed point criteria on the
forward pass is $\delta=0.1$ (and $\frac{\delta}{10}$ on the backwards) and is
based on the relative $L^2$ norm between GNN states. Standard deviations are
based on 3 seeds. If run on a single 4090 GPU one would need about 3 weeks of
\emph{total} compute.

\newcommand{\specialcell}[2][c]{%
  \begin{tabular}{@{}#1@{}}#2\end{tabular}}

\newcommand{\specialcelll}[2][l]{%
  \begin{tabular}{@{}#1@{}}#2\end{tabular}}

\begin{figure}[t]
    \centering%
    \begin{subfigure}{0.45\linewidth}
        \includegraphics[width=1\linewidth]{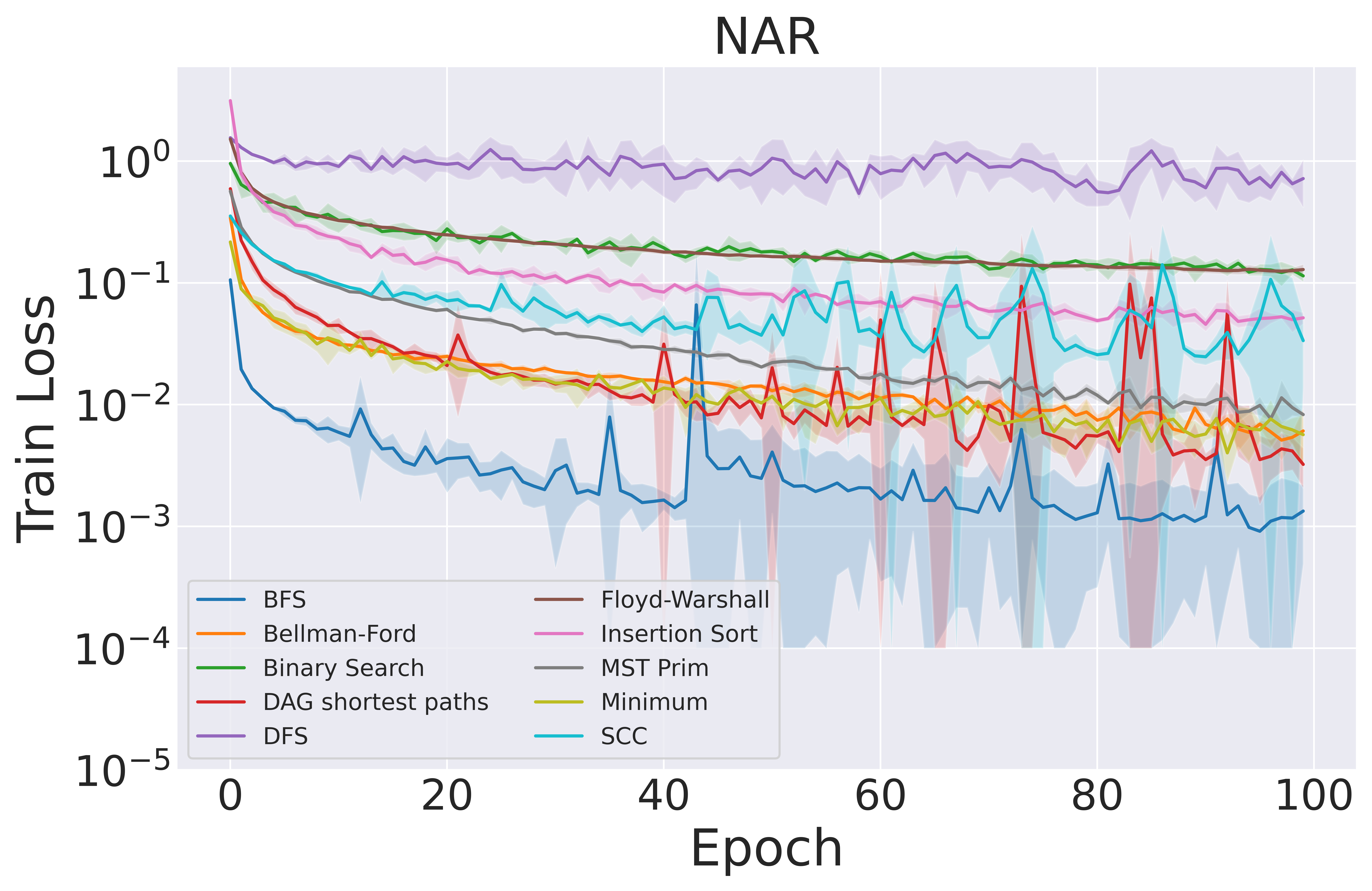}
    \end{subfigure}
    \begin{subfigure}{0.45\linewidth}
        \includegraphics[width=.884\linewidth]{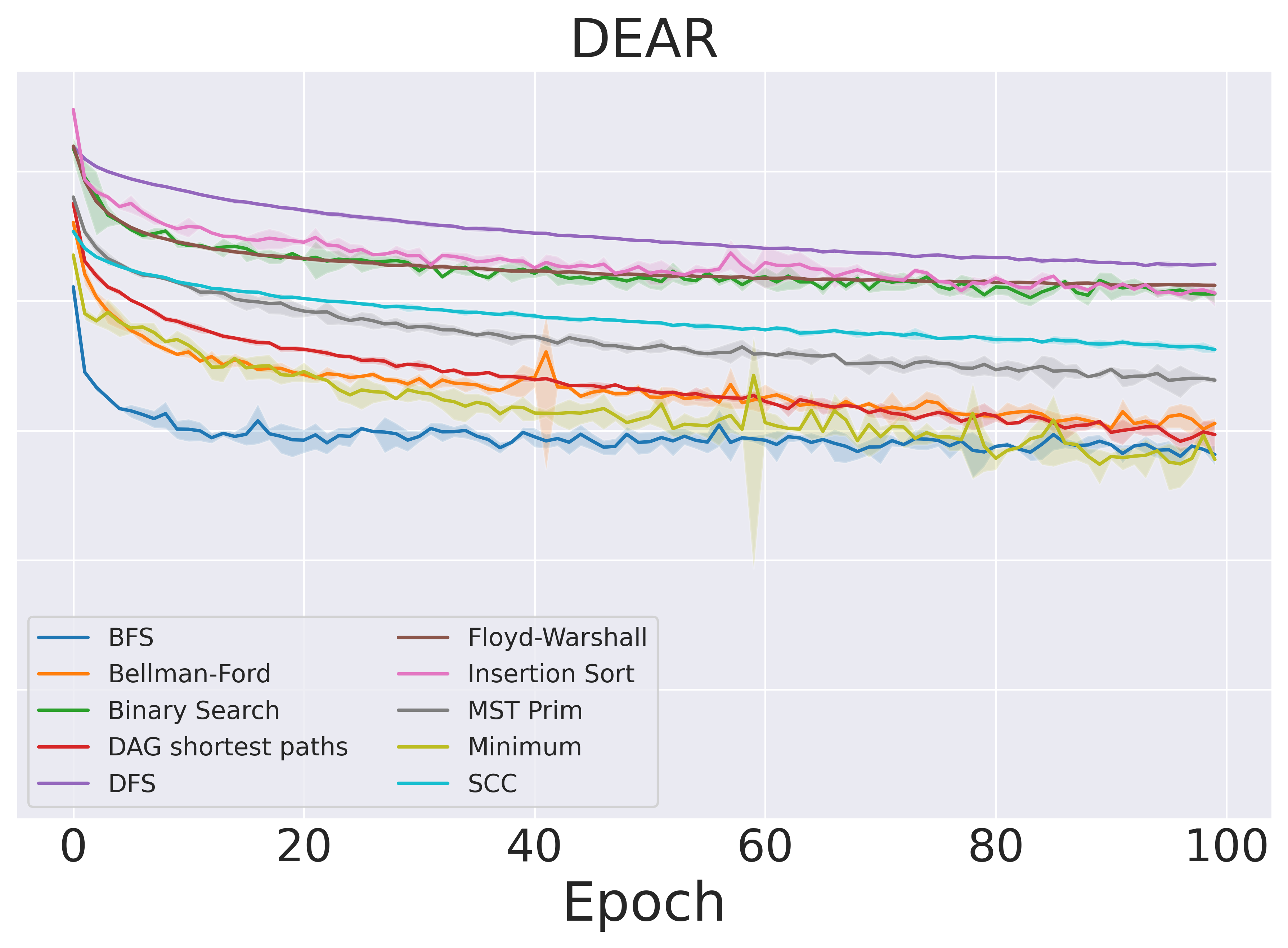}
    \end{subfigure}
    \caption{Despite converging to slightly higher train loss our models remain stable during optimisation}\label{fig:NARvsDEAR}
\end{figure}

\begin{table}[t]
\centering
\caption{%
    Test accuracy for different algorithms and models. Models with a diamond
    ($\lozenge$ or $\blacklozenge$) iterate for the correct amount of steps
    during train time (may differ between datapoints). Filled diamond
    ($\blacklozenge$) means the ground truth number of steps is also given at
    test time. LT stands for \textbf{l}earnt \textbf{t}ermination -- the model
    that uses a termination network.  For DEM
    \citep{xhonneux2024deepequilibriummodels} we leave a $-$ when no results
    are reported and we report two results for shortest path and MST as it is
    unclear to us from the main text how they differentiated between the two.
    We do not run DEAR with CGP for array tasks as they operate on
    fully-connected graphs.
}\label{tab:OHDEAR}
\footnotesize
\resizebox{\textwidth}{!}{
\begin{tabular}{lccccccccc}
\toprule
\textbf{Algorithm} & \textbf{NAR$^{\blacklozenge}$} & \specialcell{\textbf{NAR$^{\blacklozenge}$}\\(Triplet-MPNN)} & \specialcell{\textbf{NAR$^{\lozenge}$}\\(LT)} & \textbf{DEM} & \specialcell{\textbf{DEAR}\\(ours)} & \specialcell{\textbf{DEAR}\\(with CGP; ours)} \\
\midrule
\multicolumn{7}{l}{ \bf Weighted graphs}\\
\midrule
\textbf{Bellman-F.} & $97.06\% \pm 0.40$ & $97.23\%\pm 0.15$ & $95.39\%\pm 1.42$ & $96.4\%$/$78.8\%$ & $96.78\% \pm 0.43$ & $94.23\% \pm 0.59$ \\
\textbf{Floyd-W.} & $52.53\% \pm 0.98$ & $61.86\%\pm 1.57$ & $49.30\%\pm 0.53$ & - & $55.75\% \pm 2.20$ & $53.20\% \pm 2.45$ \\
\textbf{DSP} & $94.21\% \pm 1.77$ & $93.32\%\pm 1.60$ & $88.30\%\pm 1.04$ & - & $89.81\% \pm 0.14$ & $89.49\% \pm 0.17$ \\
\textbf{MST Prim} & $93.56\% \pm 0.77$ & $92.01\%\pm 1.50$ & $87.69\%\pm 1.17$ & $82.3\%$/$75.2\%$ & $88.67\% \pm 0.74$ & $86.37\% \pm 0.36$ \\
\midrule
\multicolumn{7}{l}{ \bf Unweighted graphs}\\
\midrule
\textbf{BFS} & $99.85\% \pm 0.09$ & $99.69\%\pm 0.29$ & $99.51\%\pm 0.06$ & $53.8\%$ & $98.73\% \pm 0.37$ & $98.28\% \pm 0.55$ \\
\textbf{DFS} & $16.89\% \pm 5.73$ & $31.20\%\pm 4.02$ & $29.07\%\pm 2.32$ & 5.0\% & $40.62\% \pm 0.44$ & $23.87\% \pm 2.49$ \\
\textbf{SCC} & $40.70\% \pm 1.39$ & $46.84\%\pm 1.70$ & $39.33\%\pm 1.52$ & - & $43.63\% \pm 1.19$ & $38.71\% \pm 0.45$ \\
\midrule
\multicolumn{7}{l}{\specialcelll{ \bf Arrays\\(assumes fully-connected graph)}}\\
\midrule
\textcolor{CambridgeRed}{\textbf{Search} (Binary)} & $94.67\% \pm 2.31$ & $93.33\%\pm 2.31$ & $84.33\%\pm 8.33$ & - & $59.00\% \pm 12.3$ & - \\
\textbf{Minimum} & $97.67\% \pm 5.73$ & $96.67\%\pm 2.31$ & $94.00\%\pm 2.00$ & - & $97.22\% \pm 3.82$ & - \\
{\textbf{Sort} (Ins.)} & $27.07\% \pm 10.3$ & $63.67\%\pm 39.97$ & $33.8\%\pm 12.04$ & - & $86.93\% \pm 3.87$ & - \\
\midrule
\textbf{Overall} & $71.42\%$ & $\mathbf{77.58\%}$ & $70.07\%$ & - & \underline{$75.42\%$} & - \\
\bottomrule
\end{tabular}
}
\end{table}

The performance metric we measure is the out-of-distribution accuracy, hence
the larger test instances. The definition of accuracy varies between algorithms
and is based on the specification of the algorithm itself. We refer the reader to 
\citet{velivckovic2022clrs} and CLRS-30 for corresponding accuracy metrics
definitions. The main baselines we compare against are the results
\emph{reported} by \citet{xhonneux2024deepequilibriummodels}, as no
implementation is publicly available, and a NAR architecture with a PGN
processor trained in the no-hint regime, as done by
\citet{bevilacqua2023neural}. As, logically, models that are provided the
ground-truth number of steps at test time will perform better, we also add as
additional baselines a model that always uses 64 steps at test time and a model
that has a dedicated network to decide termination
\citep{velickovic2020neural}. In order to understand how we compare to other,
architectural alignments, we also provide a comparison with a more expressive
processor (Triplet-MPNN). 

\subsection{Results} We present results for 10 key algorithms (most of
the ones used in \citet{bevilacqua2023neural}) in \autoref{tab:OHDEAR}.

\paragraph{DEARs are reasoners}
The first set of experiments aims to establish whether learning to execute
algorithms by finding the least fixed point is possible. As
\citet{xhonneux2024deepequilibriummodels} report that their models were prone
to optimisation issues, we first compared the training loss for a standard
NAR model and a DEAR model with the same neural components. The plots are
visualised in \autoref{fig:NARvsDEAR}. In line with the previous 
work, we observed that the DEAR tends to converge to a slightly higher training
loss as no algorithm's mean training loss dropped below 0.01. However, as
evident in \autoref{fig:NARvsDEAR}, we found the optimisation to be overall
stable, and the final train loss difference between NAR and DEAR was never
greater than 0.1 -- see \autoref{app:NARvsDEARindividuals}. We are unaware if
\citet{xhonneux2024deepequilibriummodels} observed the same numerical
differences, but we were overall satisfied with the convergence of DEARs.

\paragraph{Equilibrium is a useful inductive bias}
DEAR outperforms both of the above baselines achieving a 4-5\% overall score
increase, suggesting that aligning to the equilibrium property is a useful
inductive bias. Moreover, DEAR with a PGN processor is comparable to a NAR with
the more expressive Triplet-MPNN, achieving only 2\%
lower overall accuracy. This commendable achievement required no information
about the ground-truth number of steps neither at train time nor during
inference. A more detailed performance analysis follows.

On weighted graph algorithms our model performed on par with the baseline NAR
no-hint model for Bellman-Ford, it outperformed the baseline on Floyd-Warshall,
and scored slightly behind on the other two. On unweighted ones, it retained
fairly high BFS accuracy compared to DEM and it provided better scores for DFS
and Strongly Connected Components (SCC). Unfortunately, even though for this
kind of algorithms we used separate edge features for the CGP, in order to
distinguish CGP edges from input ones, CGP had a detrimental effect. We
hypothesise that algorithms like DFS and SCC need a more advanced architecture
or require different task specifications (the algorithm for SCC has a parallel
``twin''; see \citep{engelmayer2023parallel}) in order to generalise OOD. On
algorithms on arrays, we got a significant performance improvement in the
sorting task and got almost equal scores for min finding.  However, the model
underperformed by a large margin on the binary search task (in
\textcolor{CambridgeRed}{red}). This result was very concerning, so we
investigated further -- \autoref{app:BSano} showed that DEARs overfitted a lot
on the classic representation of binary search and that when the task is
designed carefully, DEARs can reach overall performance of a Triplet-MPNN NAR
architecture.

\begin{figure}[t]
    \centering%
    \begin{subfigure}{0.45\linewidth}
        \includegraphics[width=1\linewidth]{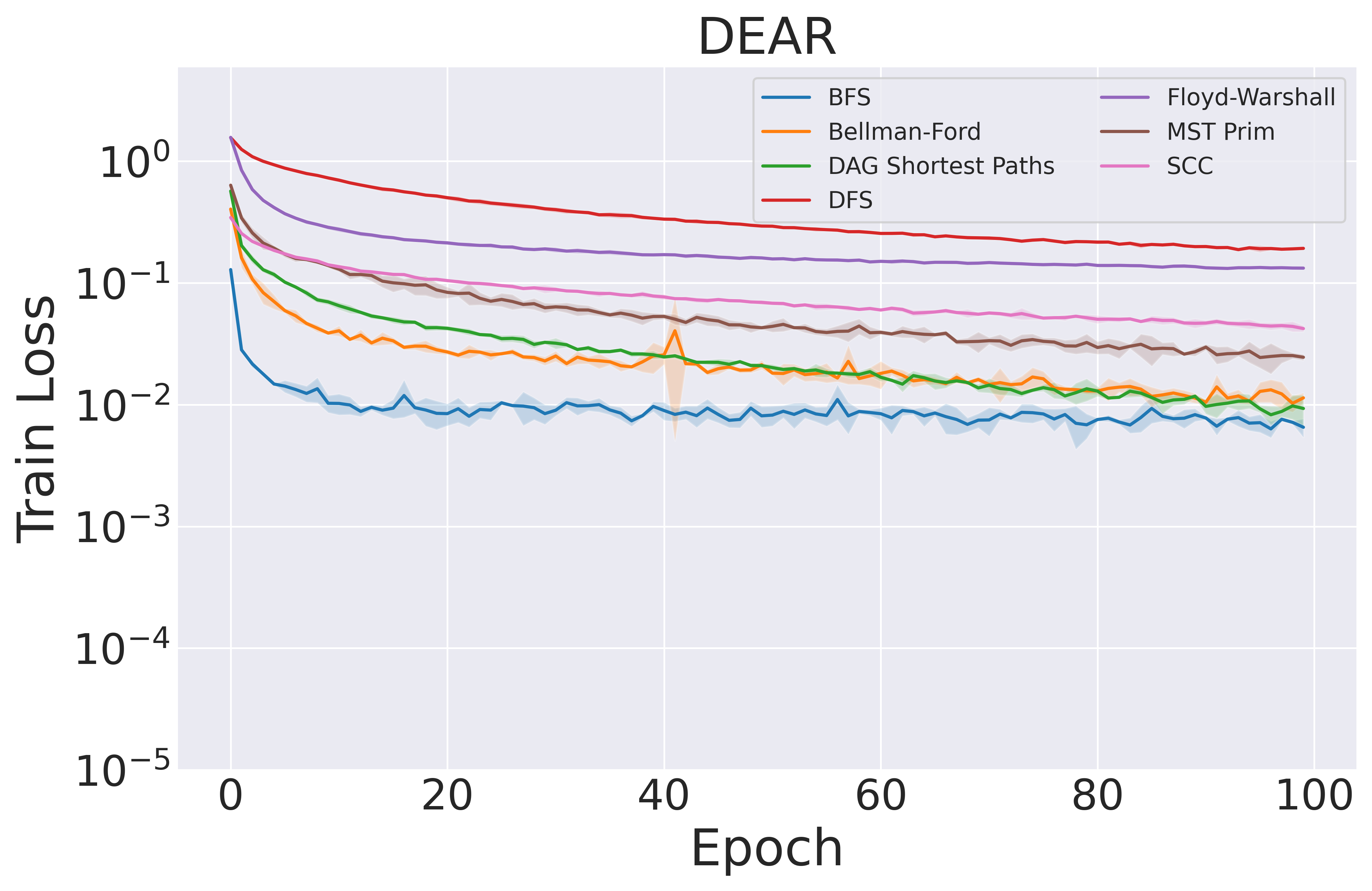}
    \end{subfigure}
    \begin{subfigure}{0.45\linewidth}
        \includegraphics[width=.884\linewidth]{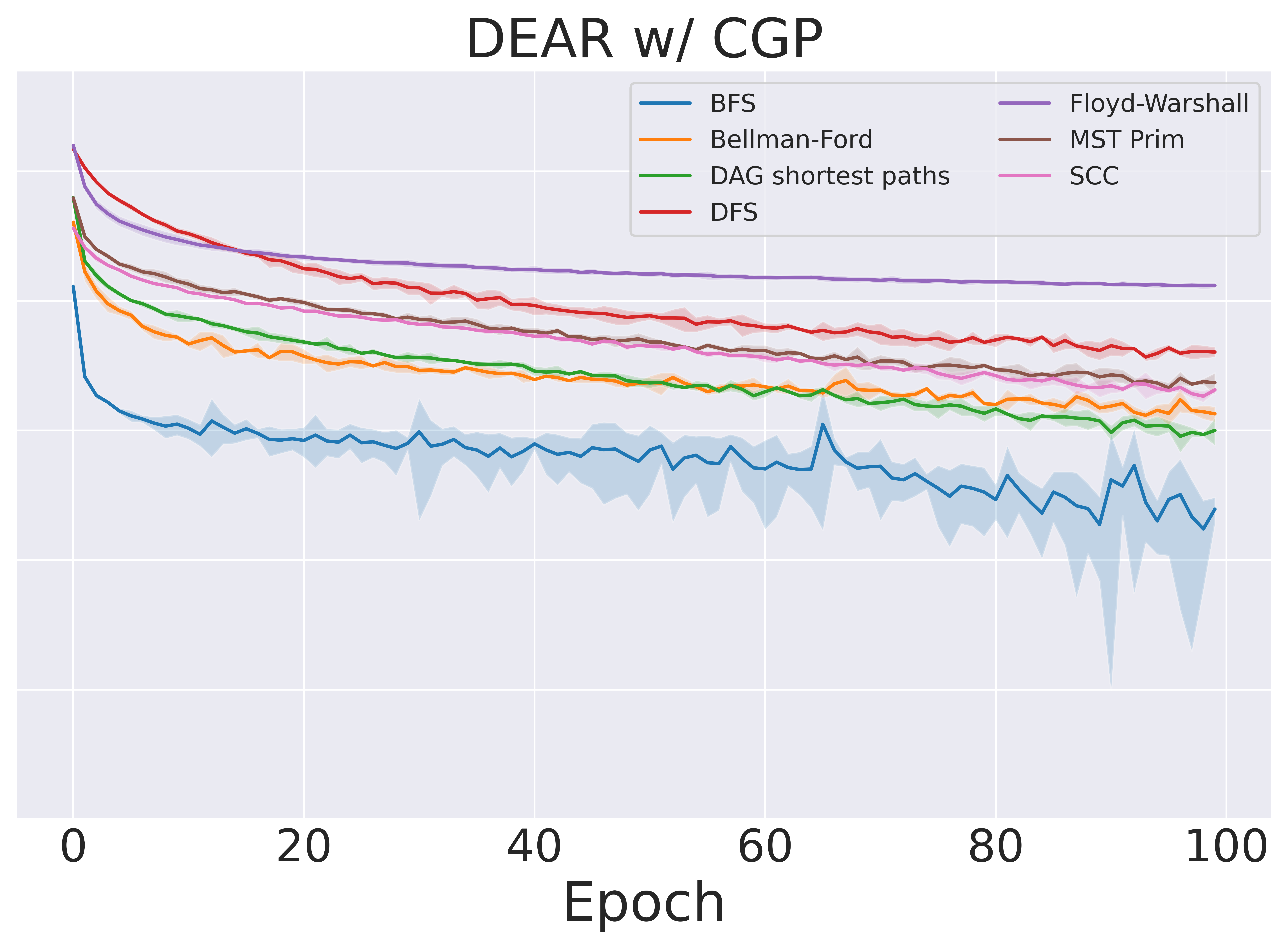}
    \end{subfigure}
    \caption{Cayely graph propagation can help with convergence}\label{fig:CGPbetterConverge}
\end{figure}
\paragraph{Effects of using CGP} Despite the slightly lower accuracies, our
experiments with CGP have not been futile. In \autoref{fig:CGPbetterConverge}, we
observe that for almost half of the algorithms CGP applies to, it had
a positive effect on the loss convergence -- 3/7 algorithms converged to at
least half an order of magnitude lower train loss. The rest did not experience
any strong negative effects of CGP. Per-algorithm plots can be found in
\autoref{app:DEARvsCGPindividuals}.  We believe that the reduced accuracies are
due to the nearest Cayley graph for the training sizes being unique and a size
of 24 nodes. Our deterministic approach of generating a fixed Cayley graph for
CGP, whose size is still distinct from test-time size leads to overfitting;
what we may observe here.  Future avenues of work may want to investigate this
by methodically removing the Cayley graph's edges, but still retaining the
desirable expansion properties \citep{banerjee2022oversquashing}, or by
exploring alternative novel graph wiring techniques
\citep{barbero2024localityaware}. However, the limitation of these proposed
approaches in comparison to CGP is that they may require \emph{dedicated preprocessing}
to scale (one of the desirable criteria set by the EGP method), therefore
providing an interesting line of future work.


\paragraph{Alignment can distill knowledge into DEARs} For evaluating our
alignment we focused on the non-CGP version of DEAR and decided to pick
algorithms where: 1) The baseline performs reasonably well (90+\% accuracy), so
as to provide good support; 2) the DEAR underperforms substantially. The
algorithms to fulfil those requirements are: DAG Shortest paths, MST Prim and
Binary Search.

\begin{figure}[t]
    \centering%
    \begin{subfigure}{0.42\linewidth}
        \includegraphics[width=1\linewidth]{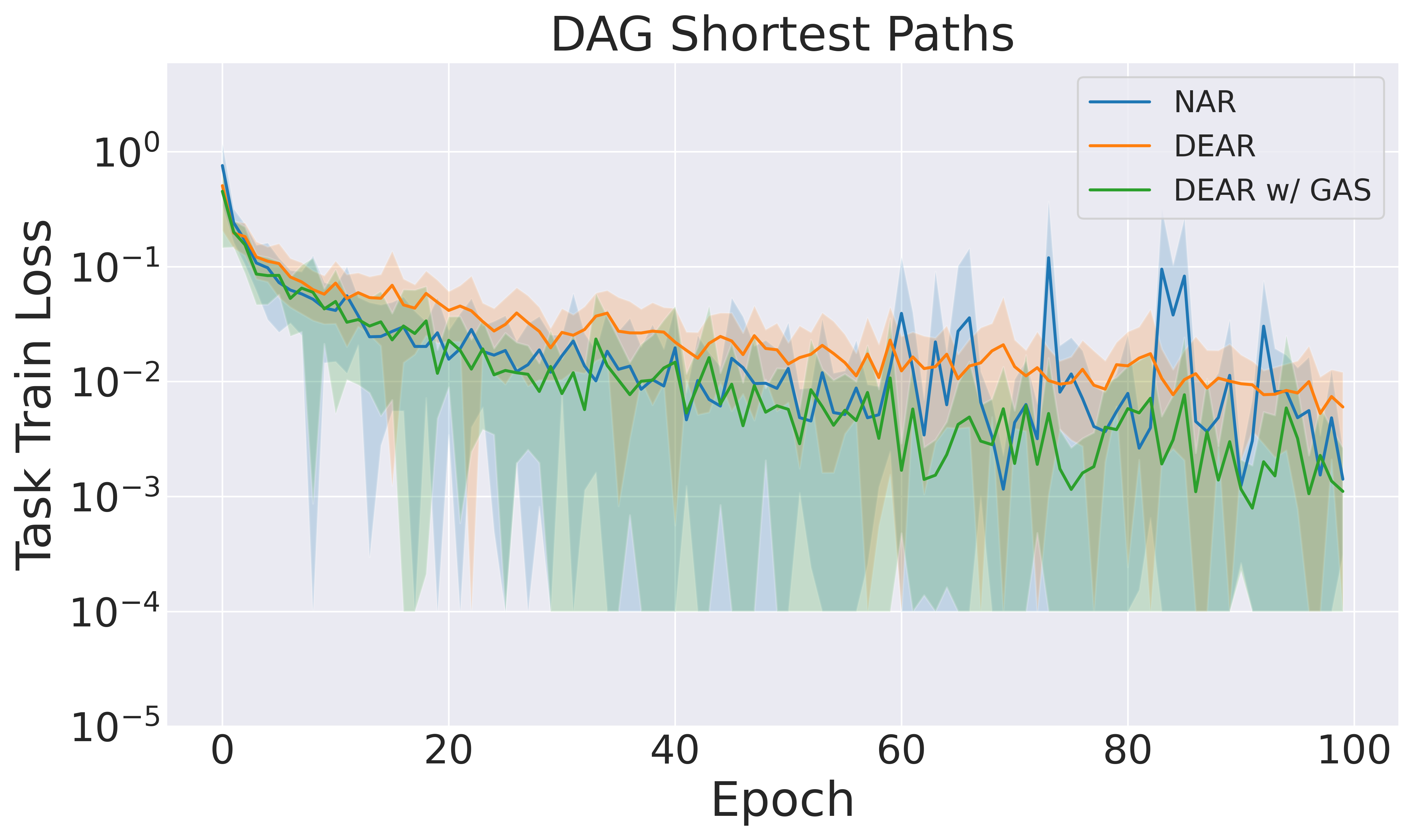}
    \end{subfigure}
    \begin{subfigure}{0.42\linewidth}
        \includegraphics[width=1\linewidth]{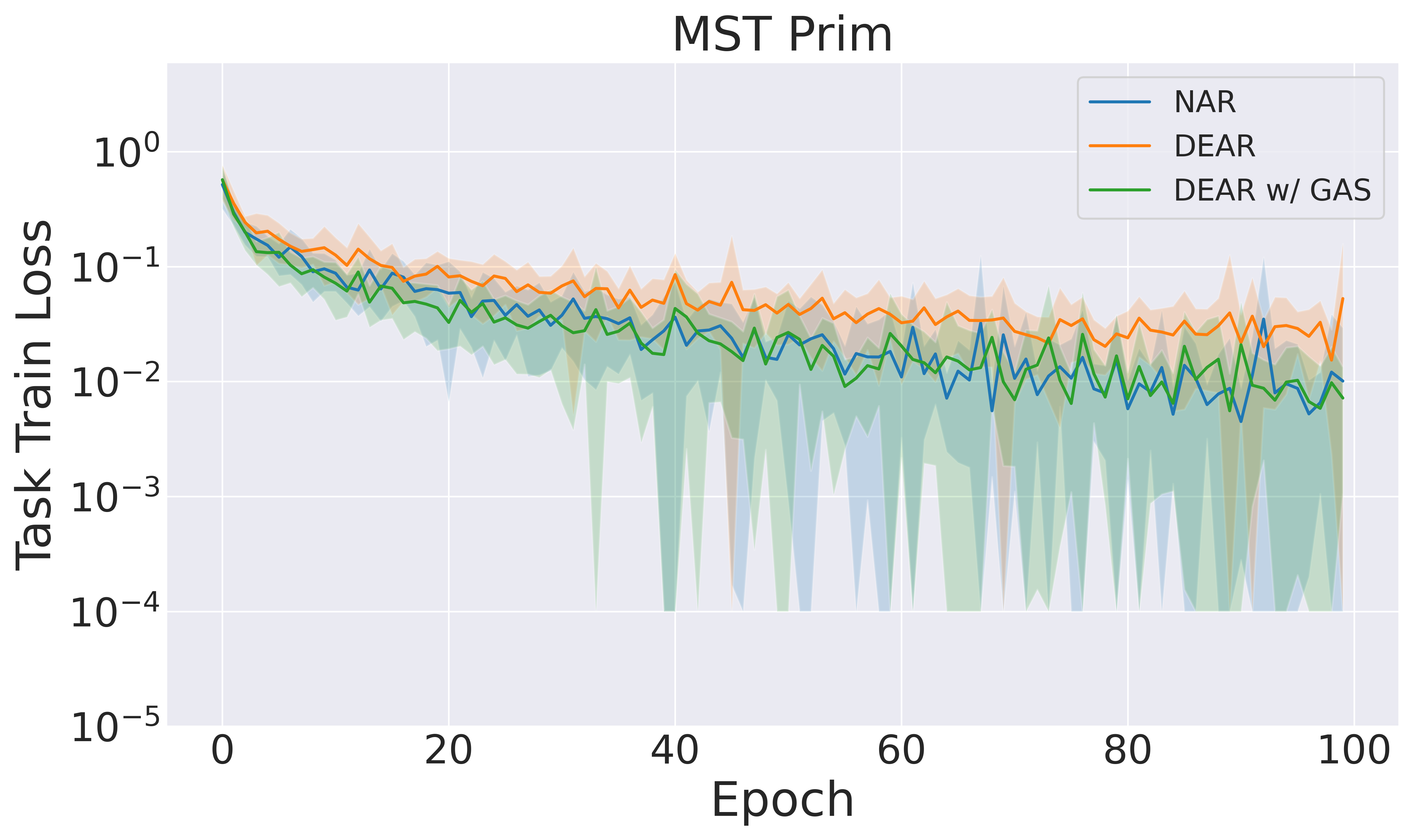}
    \end{subfigure}
    \caption{Alignment (with \textsc{granola} and stochasticity; \textbf{DEAR w/ GAS}) gives better convergence}\label{fig:threesomeAlignment}
\end{figure}

\begin{figure}[t]
    \centering%
    \begin{subfigure}{0.42\linewidth}
        \includegraphics[width=1\linewidth]{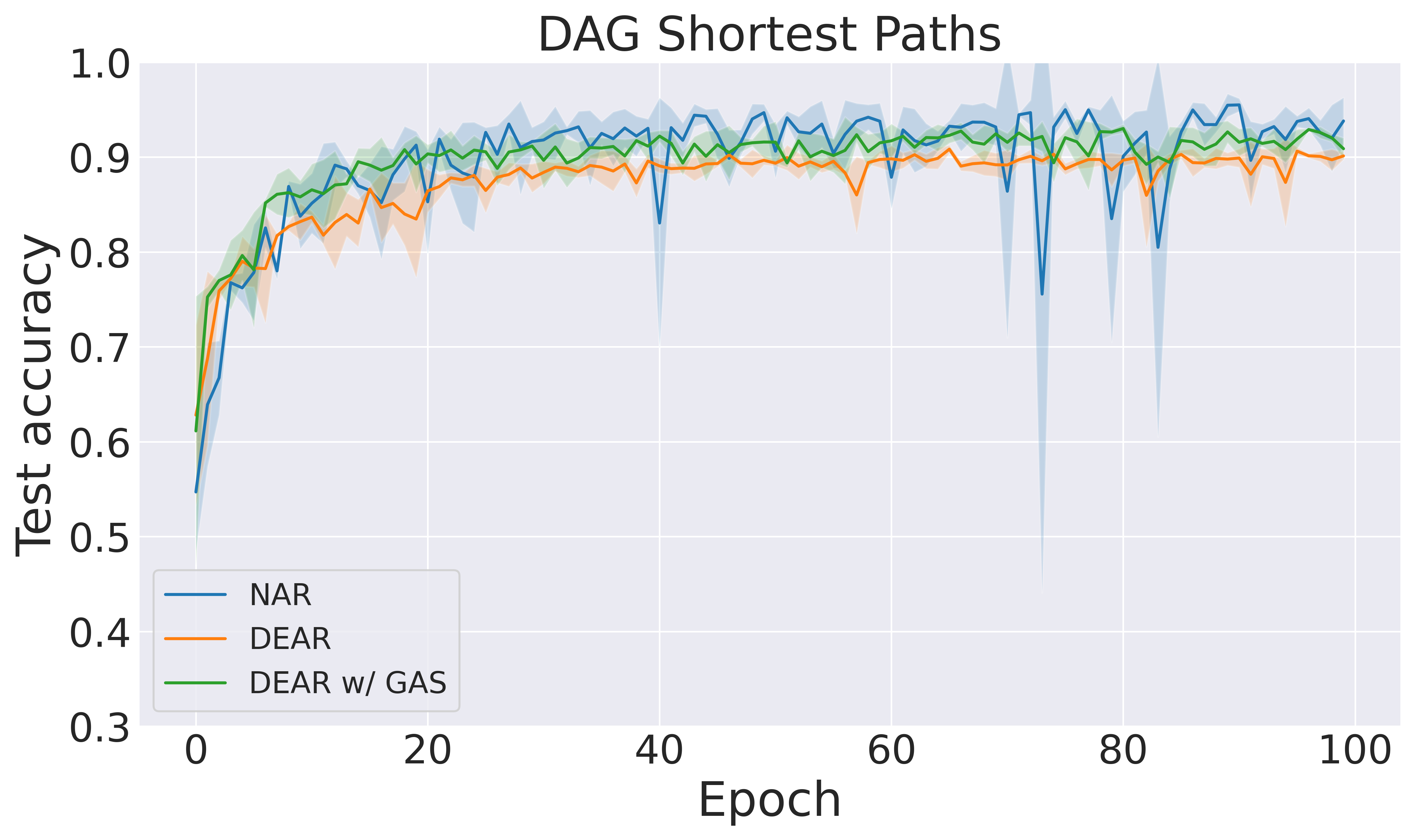}
    \end{subfigure}
    \begin{subfigure}{0.42\linewidth}
        \includegraphics[width=1\linewidth]{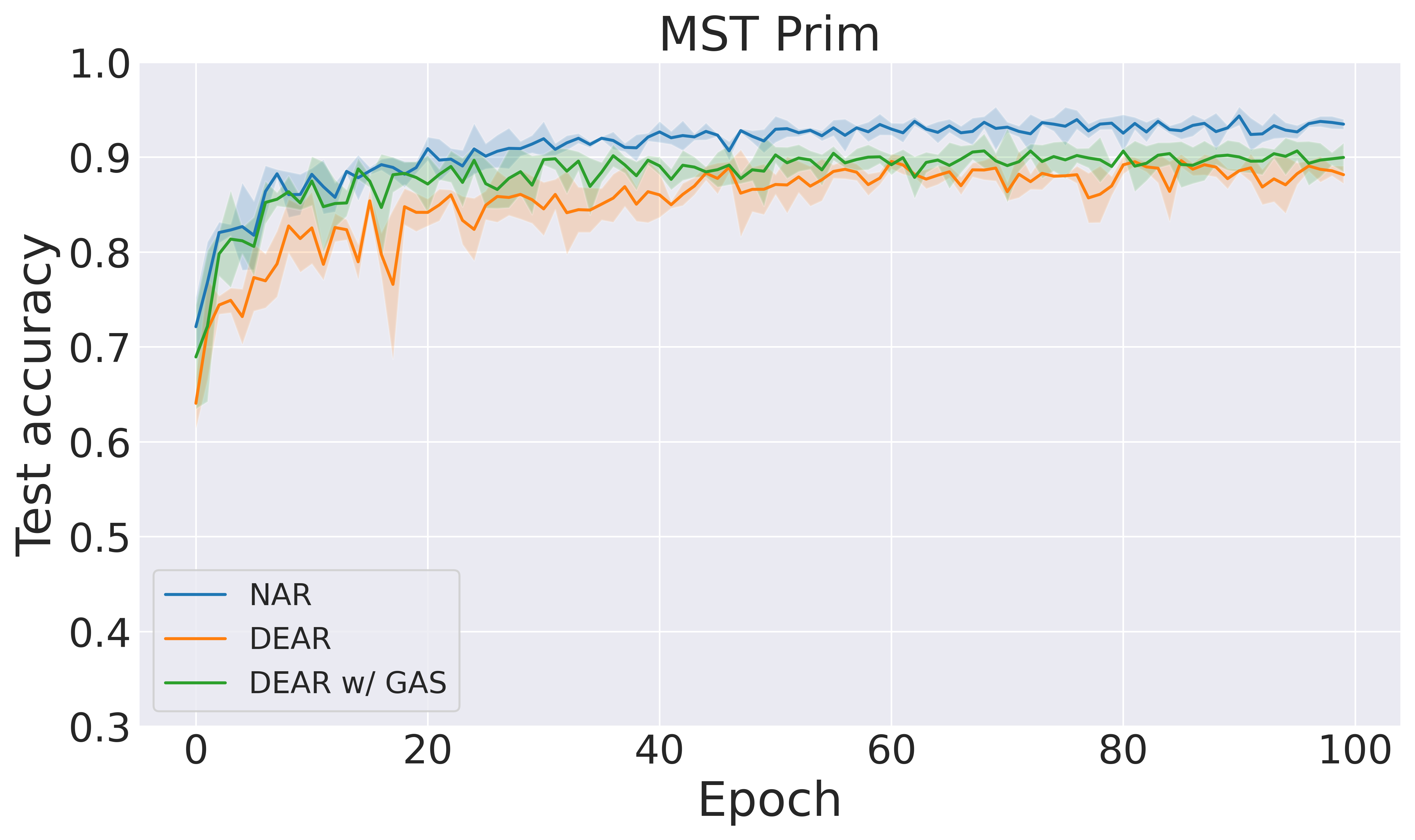}
    \end{subfigure}
    \caption{Alignment (\textbf{DEAR w/ GAS}) leads to improvements OOD}\label{fig:threesomeAlignment_pt}
\end{figure}

\begin{wraptable}{r}{0.53\textwidth}
\caption{%
Test accuracy with and without alignment.
}\label{tab:alignmebaby}
\resizebox{\linewidth}{!}{%
\begin{tabular}{lcccccc}
\toprule
&  \textbf{DSP} &  \textbf{MST-Prim} &  \textbf{Binary Search} \\
\midrule
 \textbf{NAR} &  $94.21\% \pm 1.77$ &  $93.56\% \pm 0.77$ &  $94.67\% \pm 2.31$ \\
 \textbf{DEAR} &  $89.81\% \pm 0.14$ &  $88.67\% \pm 0.74$ &  $59.00\% \pm 12.3$ \\
\specialcelll{ \textbf{DEAR}\\\scriptsize(alignment)} &  $89.65\% \pm 2.95$ &  $90.37\% \pm 1.19$ &  $77.33\% \pm 4.51$\\
\bottomrule
\end{tabular}
}
\end{wraptable}
Results are presented in \autoref{tab:alignmebaby}. At first glance, the only
algorithm that substantially improved was binary search, giving an almost 20\%
increase. The final test accuracy, however, does not represent all reality:
\autoref{fig:threesomeAlignment} shows that the task train loss (loss
excluding any regularisers) for the model with alignment decreases, compared to
no alignment and reaches similar levels as the one observed for the non-DEQ
solution. So, is it overfitting again? We argue it is not.
\autoref{fig:threesomeAlignment_pt} shows that the \emph{test (OOD)} accuracy
per epoch increases when using alignment, reaching similar accuracies to NAR
for the DAG shortest path problem and improving over plain DEAR for MST-Prim,
suggesting that choosing the right model using validation seed is hard in NAR
\citep{mahdavi2023towards}. Lastly, we would like to note that: 1) although
\textsc{granola}+stochasticity does bring benefits on its own, alignment is necessary to
narrow the gap to the NAR training loss (\autoref{app:alignftw}); 2) We never reached
perfect (0) alignment loss, suggesting better alignment techniques may further
boost performance.

\begin{table}[t]
\centering
\caption{%
    Mean inference time in seconds per sample. Measured on an RTX 4090 GPU.\@
    Up/down arrows denote improvements/deteriorations.
    $\textcolor{CambridgeBlue}{\approx}$ is used when difference is negligible.
    A double symbol is used for substantial ($5\times$) differences.
}\label{chap6:tab:performance_comparison}
\resizebox{1.05\textwidth}{!}{%
\begin{tabular}{lccccccccccc}
\toprule
 & \textbf{Bellman-F.} $\textcolor{CambridgeRed}{\uparrow}$ & \textbf{Floyd-W.} $\textcolor{CambridgeRed}{\uparrow}$ & \textbf{DSP} $\textcolor{CambridgeGreen}{\downarrow}$ & \textbf{MST Prim} $\textcolor{CambridgeGreen}{\downarrow}$ & \textbf{BFS} $\textcolor{CambridgeRed}{\uparrow}$ & \textbf{DFS} $\textcolor{CambridgeGreen}{\downarrow\downarrow}$ & \textbf{SCC} $\textcolor{CambridgeGreen}{\downarrow\downarrow}$ & \specialcell{\textbf{Search}\\(Binary)} $\textcolor{CambridgeBlue}{\approx}$ & \textbf{Minimum} $\textcolor{CambridgeGreen}{\downarrow}$ & \specialcell{\textbf{Sort}\\(Insertion)} $\textcolor{CambridgeGreen}{\downarrow\downarrow}$ \\
\midrule
\textbf{NAR}$^\blacklozenge$ & 0.0118 & 0.0916 & 0.1334 & 0.0708 & 0.0094 & 0.2440 & 0.4017 & 0.0125 & 0.0684 & 0.5680 \\
\textbf{DEAR} & 0.0215 & 0.1102 & 0.0345 & 0.0297 & 0.0137 & 0.0478 & 0.0253 & 0.0131 & 0.0174 & 0.0260 \\
\bottomrule
\end{tabular}
}
\end{table}

\begin{table}[h!]
    \caption{DEAR is architecture invariant and can also run with a Triplet-MPNN processor.}\label{chap6:tab:dear3}
\footnotesize
\resizebox{\textwidth}{!}{%
\begin{tabular}{lcccccc}
\toprule
    & \textbf{Floyd-W.} & \textbf{DFS} & \textbf{SCC} & \specialcell{\textbf{Search}\\(Parallel)} & \textbf{Sort} & \textbf{Overall} \\
\midrule
\specialcelll{\textbf{NAR$^{\blacklozenge}$}} & $61.86\% \pm 1.57$ & $31.20\% \pm 4.02$ & $\mathbf{46.84\% \pm 1.70}$ & $\mathbf{93.33\% \pm 0.58}$ & $63.67\% \pm 39.97$ & $59.18\%$ \\
\specialcelll{\textbf{DEAR}\\(ours)} & $\mathbf{62.29\% \pm 2.71}$ & $\mathbf{42.73\% \pm 2.79}$ & $45.12\% \pm 1.52$ & $87.00\% \pm 5.57$ & $\mathbf{82.34\% \pm 9.46}$ & $\mathbf{63.90\%}$\\
\bottomrule
\end{tabular}
}
\end{table}

\paragraph{DEARs are highly parallel}
DEAR is not bound to follow sequential trajectories and GNNs are more aligned
to parallel algorithms than to sequential ones \citep{engelmayer2023parallel}.
As the cost for one step of DEAR (GNN iteration + solver) is at least as high
as one step of an NAR model (GNN iteration only), we used the inference speed
of a DEAR as a measure of how parallel the final learnt algorithm is.  Results
are presented in \autoref{chap6:tab:performance_comparison}. An immediate
observation is that DEAR improves inference times across almost all algorithms.
The only ones that were executed slower are: 1) Bellman-Ford and BFS, which are
highly parallelised in the CLRS-30 implementation, so an improvement on them
was unlikely; 2) Floyd-Warshall where the difference, although present, is
marginal and we account it to the added overhead from the solver; 3) Binary
search, where performance was almost identical.  These results suggest that
although not always guaranteed (the case for searching), it is very likely that
a DEAR will learn a parallel algorithm. The most substantial improvements, in
line with our past observations in \citet{engelmayer2023parallel}, were on the
tasks of sorting and strongly-connected components.

\paragraph{DEARs are foundational}
Up until this point, DEAR was run with a PGN processor, which is a lightweight,
yet well-performant NAR processor architecture. The last set of experiments
aims to show that equilibrium reasoning is not tied to only one type of
processor/architecture. It is rather a \textbf{class of models/foundational
model} as it can natively support different types of processors. To verify
this claim, we present results with DEAR using the Triplet-MPNN architecture in
\autoref{chap6:tab:dear3}. As Triplet-MPNN is computationally expensive, we
tested algorithms for which NAR with Triplet-MPNN improves over NAR with PGN.\@
Results indeed confirm that we are not limited to a single type of processor
with DEAR, and, as expected, the best overall performance is achieved when
using DEAR with the more expressive, Triplet-MPNN, processor.

\section{Conclusion}

Our investigations with equilibrium models have shown that it is possible and
even beneficial to merge NAR and DEQs. While our models attained very
competitive performance, there are certain limitations that need to be
addressed: 1) Better algorithms for alignment can help close the gap even
further for Prim's algorithm and binary search; 2) Better model selection is
needed in order to know which DEARs would perform well OOD; 3) Graph rewiring
techniques may be needed to prevent overfitting with CGP; 4)
Algorithmic-aligned criteria for fixed-point may boost OOD generalisation for
sequential algorithms. The last point is motivated by the fact that for each
step, these algorithms update only a few nodes in the graph, keeping the rest
untouched.

\clearpage

\section*{Acknowledgements}
Dobrik Georgiev would like to acknowledge the financial support from G-Research towards covering his travel costs.

\bibliographystyle{apalike}
\bibliography{main}

\begin{thebibliography}{}

\bibitem[Alon and Yahav, 2020]{alon2020bottleneck}
Alon, U. and Yahav, E. (2020).
\newblock On the bottleneck of graph neural networks and its practical
  implications.
\newblock {\em arXiv preprint arXiv:2006.05205}.

\bibitem[Anderson, 1965]{anderson1965iterative}
Anderson, D.~G. (1965).
\newblock Iterative procedures for nonlinear integral equations.
\newblock {\em Journal of the ACM ( JACM)}.

\bibitem[Arnaiz-Rodr{\'\i}guez et~al., 2022]{arnaiz2022diffwire}
Arnaiz-Rodr{\'\i}guez, A., Begga, A., Escolano, F., and Oliver, N. (2022).
\newblock Diffwire: Inductive graph rewiring via the {L}ov{\`a}sz bound.
\newblock {\em arXiv preprint arXiv:2206.07369}.

\bibitem[Bai et~al., 2019]{bai2019deep}
Bai, S., Kolter, J.~Z., and Koltun, V. (2019).
\newblock Deep equilibrium models.
\newblock {\em Neural Information Processing Systems}.

\bibitem[Bai et~al., 2021]{bai2021stabilizing}
Bai, S., Koltun, V., and Kolter, J.~Z. (2021).
\newblock Stabilizing equilibrium models by jacobian regularization.
\newblock {\em International Conference on Machine Learning}.

\bibitem[Banerjee et~al., 2022]{banerjee2022oversquashing}
Banerjee, P.~K., Karhadkar, K., Wang, Y.~G., Alon, U., and Mont{\'u}far, G.
  (2022).
\newblock Oversquashing in gnns through the lens of information contraction and
  graph expansion.
\newblock In {\em 2022 58th Annual Allerton Conference on Communication,
  Control, and Computing (Allerton)}, pages 1--8. IEEE.

\bibitem[Barbero et~al., 2024]{barbero2024localityaware}
Barbero, F., Velingker, A., Saberi, A., Bronstein, M.~M., and Giovanni, F.~D.
  (2024).
\newblock Locality-aware graph rewiring in {GNN}s.
\newblock In {\em The Twelfth International Conference on Learning
  Representations}.

\bibitem[Bevilacqua et~al., 2023]{bevilacqua2023neural}
Bevilacqua, B., Nikiforou, K., Ibarz, B., Bica, I., Paganini, M., Blundell, C.,
  Mitrovic, J., and Velickovic, P. (2023).
\newblock Neural algorithmic reasoning with causal regularisation.
\newblock {\em International Conference on Machine Learning, {ICML} 2023, 23-29
  July 2023, Honolulu, Hawaii, {USA}}, 202:2272--2288.

\bibitem[Broyden, 1965]{broyden1965aclass}
Broyden, C.~G. (1965).
\newblock A class of methods for solving nonlinear simultaneous equations.
\newblock {\em Mathematics of Computation}, 19:577--593.

\bibitem[Cai et~al., 2023]{cai2023connection}
Cai, C., Hy, T.~S., Yu, R., and Wang, Y. (2023).
\newblock On the connection between mpnn and graph transformer.
\newblock In {\em International Conference on Machine Learning}, pages
  3408--3430. PMLR.

\bibitem[Cormen et~al., 2009]{CLRS}
Cormen, T.~H., Leiserson, C.~E., Rivest, R.~L., and Stein, C. (2009).
\newblock {\em Introduction to Algorithms, 3rd Edition}.
\newblock {MIT} Press.

\bibitem[Deac et~al., 2022]{deac2022expander}
Deac, A., Lackenby, M., and Veli{\v{c}}kovi{\'c}, P. (2022).
\newblock Expander graph propagation.
\newblock In {\em Learning on Graphs Conference}, pages 38--1. PMLR.

\bibitem[Dudzik and Veli{\v{c}}kovi{\'c}, 2022]{dudzik2022graph}
Dudzik, A.~J. and Veli{\v{c}}kovi{\'c}, P. (2022).
\newblock Graph neural networks are dynamic programmers.
\newblock {\em Advances in Neural Information Processing Systems},
  35:20635--20647.

\bibitem[Dudzik et~al., 2024]{dudzik2024async}
Dudzik, A.~J., von Glehn, T., Pascanu, R., and Veli{\v c}kovi{\' c}, P. (2024).
\newblock Asynchronous algorithmic alignment with cocycles.
\newblock In Villar, S. and Chamberlain, B., editors, {\em Proceedings of the
  Second Learning on Graphs Conference}, volume 231 of {\em Proceedings of
  Machine Learning Research}, pages 3:1--3:17. PMLR.

\bibitem[Eliasof et~al., 2024]{eliasof2024granola}
Eliasof, M., Bevilacqua, B., Schönlieb, C.-B., and Maron, H. (2024).
\newblock Granola: Adaptive normalization for graph neural networks.
\newblock {\em arXiv preprint arXiv: 2404.13344}.

\bibitem[Engelmayer et~al., 2023]{engelmayer2023parallel}
Engelmayer, V., Georgiev, D.~G., and Veli{\v{c}}kovi{\'c}, P. (2023).
\newblock Parallel algorithms align with neural execution.
\newblock In {\em The Second Learning on Graphs Conference}.

\bibitem[Erdős et~al., 1960]{erdos1960evolution}
Erdős, P., R{\'e}nyi, A., et~al. (1960).
\newblock On the evolution of random graphs.
\newblock {\em Publ. Math. Inst. Hung. Acad. Sci}, 5(1):17--60.

\bibitem[Fiore, 2324]{fioredenotational}
Fiore, M. (2023/24).
\newblock {\em Denotational Semantics. Lecture notes for Part II of the
  Computer Science Tripos}.
\newblock University of Cambridge.

\bibitem[Geng and Kolter, 2023]{torchdeq}
Geng, Z. and Kolter, J.~Z. (2023).
\newblock Torchdeq: A library for deep equilibrium models.
\newblock \url{https://github.com/locuslab/torchdeq}.

\bibitem[Ghaoui et~al., 2019]{ghaoui2019implicit}
Ghaoui, L., Gu, F., Travacca, B., Askari, A., and Tsai, A.~Y. (2019).
\newblock Implicit deep learning.
\newblock {\em SIAM Journal on Mathematics of Data Science}.

\bibitem[Gilmer et~al., 2017]{gilmer2017neural}
Gilmer, J., Schoenholz, S.~S., Riley, P.~F., Vinyals, O., and Dahl, G.~E.
  (2017).
\newblock Neural message passing for quantum chemistry.
\newblock In {\em International conference on machine learning}, pages
  1263--1272. PMLR.

\bibitem[Giovanni et~al., 2023]{giovanni2023oversquashing}
Giovanni, F.~D., Rusch, T.~K., Bronstein, M.~M., Deac, A., Lackenby, M.,
  Mishra, S., and Veličković, P. (2023).
\newblock How does over-squashing affect the power of gnns?
\newblock {\em arXiv preprint arXiv: 2306.03589}.

\bibitem[Gunter, 1992]{gunter1992semantics}
Gunter, C.~A. (1992).
\newblock {\em Semantics of programming languages: structures and techniques}.
\newblock MIT press.

\bibitem[Hamrick et~al., 2018]{hamrick2018relational}
Hamrick, J.~B., Allen, K.~R., Bapst, V., Zhu, T., McKee, K.~R., Tenenbaum,
  J.~B., and Battaglia, P.~W. (2018).
\newblock Relational inductive bias for physical construction in humans and
  machines.
\newblock {\em arXiv preprint arXiv:1806.01203}.

\bibitem[Harris and Ross, 1955]{harris1955fundamentals}
Harris, T. and Ross, F. (1955).
\newblock Fundamentals of a method for evaluating rail net capacities.
\newblock Technical report.

\bibitem[Hu et~al., 2021]{hu2021ogb}
Hu, W., Fey, M., Ren, H., Nakata, M., Dong, Y., and Leskovec, J. (2021).
\newblock Ogb-lsc: A large-scale challenge for machine learning on graphs.
\newblock {\em arXiv preprint arXiv:2103.09430}.

\bibitem[Hu et~al., 2020]{hu2020open}
Hu, W., Fey, M., Zitnik, M., Dong, Y., Ren, H., Liu, B., Catasta, M., and
  Leskovec, J. (2020).
\newblock Open graph benchmark: Datasets for machine learning on graphs.
\newblock {\em Advances in neural information processing systems},
  33:22118--22133.

\bibitem[Hwang et~al., 2022]{hwang2022analysis}
Hwang, E., Thost, V., Dasgupta, S.~S., and Ma, T. (2022).
\newblock An analysis of virtual nodes in graph neural networks for link
  prediction.
\newblock In {\em The First Learning on Graphs Conference}.

\bibitem[Ibarz et~al., 2022]{ibarz2022generalist}
Ibarz, B., Kurin, V., Papamakarios, G., Nikiforou, K., Bennani, M.,
  Csord{\'{a}}s, R., Dudzik, A.~J., Bosnjak, M., Vitvitskyi, A., Rubanova, Y.,
  Deac, A., Bevilacqua, B., Ganin, Y., Blundell, C., and Velickovic, P. (2022).
\newblock A generalist neural algorithmic learner.
\newblock In Rieck, B. and Pascanu, R., editors, {\em Learning on Graphs
  Conference, LoG 2022, 9-12 December 2022, Virtual Event}, volume 198 of {\em
  Proceedings of Machine Learning Research}, page~2. {PMLR}.

\bibitem[Karhadkar et~al., 2022]{karhadkar2022fosr}
Karhadkar, K., Banerjee, P.~K., and Mont{\'u}far, G. (2022).
\newblock Fosr: First-order spectral rewiring for addressing oversquashing in
  gnns.
\newblock {\em arXiv preprint arXiv:2210.11790}.

\bibitem[Kingma and Ba, 2015]{kingma2015adam}
Kingma, D.~P. and Ba, J. (2015).
\newblock Adam: {A} method for stochastic optimization.
\newblock In Bengio, Y. and LeCun, Y., editors, {\em 3rd International
  Conference on Learning Representations, {ICLR} 2015, San Diego, CA, USA, May
  7-9, 2015, Conference Track Proceedings}.

\bibitem[Kolter et~al., 2023]{kolter2023deep}
Kolter, Z., Duvenaud, D., and Johnson, M. (2023).
\newblock Deep equilibrium models (deq) tutorial.
\newblock \url{https://implicit-layers-tutorial.org/deep_equilibrium_models/}.
\newblock Accessed: 2024-08-27.

\bibitem[Liu et~al., 2024]{liu2024scalable}
Liu, J., Hooi, B., Kawaguchi, K., Wang, Y., Dong, C., and Xiao, X. (2024).
\newblock Scalable and effective implicit graph neural networks on large
  graphs.
\newblock In {\em The Twelfth International Conference on Learning
  Representations}.

\bibitem[Mahdavi et~al., 2023]{mahdavi2023towards}
Mahdavi, S., Swersky, K., Kipf, T., Hashemi, M., Thrampoulidis, C., and Liao,
  R. (2023).
\newblock Towards better out-of-distribution generalization of neural
  algorithmic reasoning tasks.
\newblock {\em Transactions on Machine Learning Research}.

\bibitem[Mohar, 1991]{mohar1991eigenvalues}
Mohar, B. (1991).
\newblock Eigenvalues, diameter, and mean distance in graphs.
\newblock {\em Graphs and combinatorics}, 7(1):53--64.

\bibitem[Paszke et~al., 2019]{paszke2019pytorch}
Paszke, A., Gross, S., Massa, F., Lerer, A., Bradbury, J., Chanan, G., Killeen,
  T., Lin, Z., Gimelshein, N., Antiga, L., et~al. (2019).
\newblock Pytorch: An imperative style, high-performance deep learning library.
\newblock {\em arXiv preprint arXiv:1912.01703}.

\bibitem[Rodionov and Prokhorenkova, 2023]{rodionov2023neural}
Rodionov, G. and Prokhorenkova, L. (2023).
\newblock Neural algorithmic reasoning without intermediate supervision.
\newblock {\em arXiv preprint arXiv:2306.13411}.

\bibitem[Scott, 1982]{scott1982domains}
Scott, D.~S. (1982).
\newblock Domains for denotational semantics.
\newblock In {\em Automata, Languages and Programming: Ninth Colloquium Aarhus,
  Denmark, July 12--16, 1982 9}, pages 577--610. Springer.

\bibitem[Scott and Strachey, 1971]{scott1971toward}
Scott, D.~S. and Strachey, C. (1971).
\newblock {\em Toward a mathematical semantics for computer languages},
  volume~1.
\newblock Oxford University Computing Laboratory, Programming Research Group
  Oxford.

\bibitem[Sharir, 1981]{sharir1981strong}
Sharir, M. (1981).
\newblock A strong-connectivity algorithm and its applications in data flow
  analysis.
\newblock {\em Computers \& Mathematics with Applications}, 7(1):67--72.

\bibitem[Shirzad et~al., 2023]{shirzad2023exphormer}
Shirzad, H., Velingker, A., Venkatachalam, B., Sutherland, D.~J., and Sinop,
  A.~K. (2023).
\newblock Exphormer: Sparse transformers for graphs.
\newblock In {\em International Conference on Machine Learning}, pages
  31613--31632. PMLR.

\bibitem[Tang et~al., 2020]{tang2020towards}
Tang, H., Huang, Z., Gu, J., Lu, B.-L., and Su, H. (2020).
\newblock Towards scale-invariant graph-related problem solving by iterative
  homogeneous gnns.
\newblock {\em Advances in Neural Information Processing Systems}, 33.

\bibitem[Tarski, 1955]{tarski1955lattice}
Tarski, A. (1955).
\newblock A lattice-theoretical fixpoint theorem and its applications.

\bibitem[Topping et~al., 2021]{topping2021understanding}
Topping, J., Di~Giovanni, F., Chamberlain, B.~P., Dong, X., and Bronstein,
  M.~M. (2021).
\newblock Understanding over-squashing and bottlenecks on graphs via curvature.
\newblock {\em arXiv preprint arXiv:2111.14522}.

\bibitem[Veli{\v{c}}kovi{\'c} et~al., 2022]{velivckovic2022clrs}
Veli{\v{c}}kovi{\'c}, P., Badia, A.~P., Budden, D., Pascanu, R., Banino, A.,
  Dashevskiy, M., Hadsell, R., and Blundell, C. (2022).
\newblock The clrs algorithmic reasoning benchmark.
\newblock In {\em International Conference on Machine Learning}, pages
  22084--22102. PMLR.

\bibitem[Velickovic and Blundell, 2021a]{velickovic2021neural}
Velickovic, P. and Blundell, C. (2021a).
\newblock Neural algorithmic reasoning.
\newblock {\em Patterns}, 2(7):100273.

\bibitem[Velickovic and Blundell, 2021b]{NARBlueprint}
Velickovic, P. and Blundell, C. (2021b).
\newblock Neural algorithmic reasoning.
\newblock {\em Patterns}, 2(7):100273.

\bibitem[Veli{\v{c}}kovi{\'c} et~al., 2020]{velickovic2020pointer}
Veli{\v{c}}kovi{\'c}, P., Buesing, L., Overlan, M., Pascanu, R., Vinyals, O.,
  and Blundell, C. (2020).
\newblock Pointer graph networks.
\newblock {\em Advances in Neural Information Processing Systems},
  33:2232--2244.

\bibitem[Veli\v{c}kovi\'{c} et~al., 2020]{velickovic2020neural}
Veli\v{c}kovi\'{c}, P., Ying, R., Padovano, M., Hadsell, R., and Blundell, C.
  (2020).
\newblock Neural execution of graph algorithms.
\newblock In {\em 8th International Conference on Learning Representations,
  {ICLR} 2020, Addis Ababa, Ethiopia, April 26-30, 2020}. OpenReview.net.

\bibitem[Veličković, 2023]{velickovic2023NARGradient}
Veličković, P. (2023).
\newblock Neural algorithmic reasoning.
\newblock {\em The Gradient}.

\bibitem[Wilson et~al., 2024]{wilson2024cayley}
Wilson, J., Bechler-Speicher, M., and Veli{\v{c}}kovi{\'c}, P. (2024).
\newblock Cayley graph propagation.
\newblock {\em arXiv preprint arXiv:2410.03424}.

\bibitem[Winskel, 1993]{winskel1993formal}
Winskel, G. (1993).
\newblock {\em The formal semantics of programming languages: an introduction}.
\newblock MIT press.

\bibitem[Xhonneux et~al., 2024]{xhonneux2024deepequilibriummodels}
Xhonneux, S., He, Y., Deac, A., Tang, J., and Gidel, G. (2024).
\newblock Deep equilibrium models for algorithmic reasoning.
\newblock In {\em ICLR Blogposts 2024}.
\newblock https://iclr-blogposts.github.io/2024/blog/deqalg-reasoning/.

\bibitem[Xu et~al., 2018]{xu2018representation}
Xu, K., Li, C., Tian, Y., Sonobe, T., Kawarabayashi, K.-i., and Jegelka, S.
  (2018).
\newblock Representation learning on graphs with jumping knowledge networks.
\newblock In {\em International conference on machine learning}, pages
  5453--5462. PMLR.

\bibitem[Xu et~al., 2020]{xu2020what}
Xu, K., Li, J., Zhang, M., Du, S.~S., ichi Kawarabayashi, K., and Jegelka, S.
  (2020).
\newblock What can neural networks reason about?
\newblock In {\em International Conference on Learning Representations}.

\end{thebibliography}

\newpage 
\appendix

\section{IMP: Definitions and examples}\label{app:ebigo2}

Anything highlighted in \ctexttt{blue} below, is part of the \textbf{IMP} language. Subexpressions, to avoid confusion, are not coloured.

\textbf{IMP} consists of:
\begin{itemize}
    \item \emph{numbers} -- 1, -20, 13930
    \item \emph{locations} -- \ctexttt{AVariableName}, \ctexttt{AnotherVar}, \ctexttt{ArrayName}$\lbrack$11$\rbrack$
    \item \emph{arithmetic expressions} -- \ctexttt{(5+4)*3}, but also
        \ctexttt{A*55}. Note how variables can be parts of arithmetic expressions.
    \item \emph{boolean expressions} -- \ctexttt{true}, \ctexttt{false}, but
        also \ctexttt{X==0}, \ctexttt{3*A<B} and
        \ctexttt{X==0}\textcolor{NavyBlue}{\ $\land$ }\ctexttt{3*A<B}. Note how
        boolean expressions can be made by using variables or using boolean logic
        on sub-expressions.
    \item \emph{commands} -- Commands can be one of:
        \begin{itemize}
            \item \ctexttt{skip}, which is a no-op
            \item \ctexttt{X:=a}, where X is
        assigned the value of arithmetic expression $a$. An example $a$ is \ctexttt{3*B+C}
            \item \ctexttt{if}\ $b$\ \ctexttt{then}\ $c_0$\ \ctexttt{else}\ $c_1$ where $b$ is a boolean
                expression and $c_0$, $c_1$ are commands. For example
                \ctexttt{if A < B then C:=0 else C:=A-B} which sets \texttt{C}
                to the difference of A and B, if it is positive \item
                \ctexttt{(}$c_0$\ctexttt{;}$c_1$\ctexttt{)}
            \item $\ctexttt{while}\ b\ \ctexttt{do}\ c$ which is a while loop
                repeating command $c$ as long as boolean expression $b$.
                A (classic) example is \ctexttt{(Y:=1; while  X > 0 do (Y:=X*Y;
                X:=X-1))}, which finds the factorial of \texttt{X} and saves it
                in \texttt{Y}.
        \end{itemize}
\end{itemize}

\section{Fixed point of a while loop -- example}\label{app:ebigo}

Below, we will denote the state as $[A\mapsto a, B\mapsto b, \dots]$. It means
that the value of variable $A$ is $a$ and so on.

Consider the facorial example from above removing the explicit set of
\texttt{Y} to 1. To find the denotation $\llbracket w \rrbracket
= \llbracket$\ctexttt{while X > 0 do (Y:=X*Y; X:=X-1)}$\rrbracket$, we first
define our $f_{b,c}$

\begin{align*}
    f_{b,c}(w)([X\mapsto x, Y\mapsto y]) =
    \begin{cases}
        {w}\left([X\mapsto x-1, Y\mapsto y*x]\right) & \text{if } X > 0\\
        [X\mapsto x, Y\mapsto y] & \text{otherwise}
    \end{cases}
\end{align*}
The equivalent definition if we were to keep the lambdas from the original text is
\begin{align*}
    f_{b,c} = \lambda w \in (State \rightharpoonup State). \lambda s \in State.\
    \begin{cases}
        {w}\left([X\mapsto x-1, Y\mapsto y*x]\right) & \text{if } X > 0\\
        [X\mapsto x, Y\mapsto y] & \text{otherwise}
    \end{cases}
\end{align*}
but we will work with the first definition as it is more compact.

The approximations of $f^n_{b,c}$ starting from $f^0_{b,c}=\bot$ are:
\begin{align*}
    f^1_{b,c} &= f_{b,c}(f^0_{b,c})([X\mapsto x, Y \mapsto y]) &=&&\\
              &= f_{b,c}(\bot)([X\mapsto x, Y \mapsto y]) &=&
                    \begin{cases}
                        {\bot}\left([X\mapsto x-1, Y\mapsto y*x]\right) & \text{if } x > 0\\
                        [X\mapsto x, Y\mapsto y] & \text{otherwise}
                    \end{cases}\\
              &                                          & =& \begin{cases}
                        \text{undefined} & \text{if } x > 0\\
                        [X\mapsto x, Y\mapsto y] & \text{otherwise}
                    \end{cases}\\
    f^2_{b,c} &= f_{b,c}(f^1_{b,c})([X\mapsto x, Y \mapsto y]) &=&
                    \begin{cases}
                        {f^1_{b,c}}\left([X\mapsto x-1, Y\mapsto y*x]\right) & \text{if } x > 0\\
                        [X\mapsto x, Y\mapsto y] & \text{otherwise}
                    \end{cases}\\
              &                                          & =& \begin{cases}
                        \text{undefined} & \text{if } x-1 > 0\\
                        [X\mapsto x-1, Y\mapsto y\ast x] & \text{if } x-1 = 0\\
                        [X\mapsto x, Y\mapsto y] & \text{if } x \leq 0
                    \end{cases}\\
              &                                          & =& \begin{cases}
                        \text{undefined} & \text{if } x > 1\\
                        [X\mapsto 0, Y\mapsto y] & \text{if } x = 1\\
                        [X\mapsto x, Y\mapsto y] & \text{if } x \leq 0
                    \end{cases}\\
    f^3_{b,c} &= f_{b,c}(f^2_{b,c})([X\mapsto x, Y \mapsto y]) &=&
                    \begin{cases}
                        {f^2_{b,c}}\left([X\mapsto x-1, Y\mapsto y*x]\right) & \text{if } x > 0\\
                        [X\mapsto x, Y\mapsto y] & \text{if } x \leq 0
                    \end{cases}\\
              &                                          & =& \begin{cases}
                        \text{undefined} & \text{if } x-1 > 1\\
                        [X\mapsto 0, Y\mapsto y\ast x] & \text{if } x-1 = 1\\
                        [X\mapsto x-1, Y\mapsto y] & \text{if } x-1 \leq 0\\
                        [X\mapsto x, Y\mapsto y] & \text{if } x \leq 0\\
                    \end{cases}\\
              &                                          & =& \begin{cases}
                        \text{undefined} & \text{if } x > 2\\
                        [X\mapsto 0, Y\mapsto y\ast 2] & \text{if } x = 2\\
                        [X\mapsto 0, Y\mapsto y] & \text{if } x = 1\\
                        [X\mapsto x, Y\mapsto y] & \text{if } x \leq 0\\
                    \end{cases}\\
              &&\vdots\\
\end{align*}
which for $n$ is:
\begin{align*}
    f^n_{b,c} &= 
                    \begin{cases}
                        \text{undefined} & \text{if } x \geq n\\
                        [X\mapsto 0, Y\mapsto y \ast (x!)] & \text{if } 0 < x < n\\
                        [X\mapsto x, Y\mapsto y] & \text{if } x \leq 0\\
                    \end{cases}
\end{align*}
The sequence obeys $f^0_{b,c}\sqsubseteq f^1_{b,c} \sqsubseteq
f^2_{b,c}\sqsubseteq \dots \sqsubseteq f^n_{b,c}\sqsubseteq \dots$ (we can see
that whenever $f^{n-1}_{b,c}$ is defined it agrees with $f^{n}_{b,c}$) and
$f_{b,c}$ is monotonic ($f^k_{b,c} \sqsubseteq f^l_{b,c} \implies
f_{b,c}(f^k_{b,c}) \sqsubseteq f_{b,c}(f^l_{b,c})$). 

For a given $X=x$, $f^{x+1}_{b,c}=f^{x+2}_{b,c}=\dots$ . The fixed point is the
lub of the whole sequence is therefore: 
\begin{align*}
    f^\infty_{b,c}=\bigsqcup_{n\geq0}f^n_{b,c}(\bot)=\begin{cases}
                        [X\mapsto 0, Y\mapsto y \ast (x!)] & \text{if } x > 0\\
                        [X\mapsto x, Y\mapsto y] & \text{if } x \leq 0
                    \end{cases}
\end{align*}

By a similar analysis, it is not hard show that the denotation of
$\llbracket\ctexttt{while true do skip}\rrbracket$ will be
undefined.\footnote{omitted in this text -- see 
\citet{winskel1993formal}}

\definecolor{codegreen}{rgb}{0,0.6,0}
\definecolor{codegray}{rgb}{0.5,0.5,0.5}
\definecolor{codepurple}{rgb}{0.58,0,0.82}
\definecolor{backcolour}{rgb}{0.95,0.95,0.92}

\lstdefinestyle{mystyle}{
    backgroundcolor=\color{backcolour},   
    commentstyle=\color{codegreen},
    keywordstyle=\color{magenta},
    numberstyle=\tiny\color{codegray},
    stringstyle=\color{codepurple},
    basicstyle=\ttfamily\scriptsize,
    breakatwhitespace=false,         
    breaklines=true,                 
    captionpos=b,                    
    keepspaces=true,                 
    numbers=left,                    
    numbersep=5pt,                  
    showspaces=false,                
    showstringspaces=false,
    showtabs=false,                  
    tabsize=2
}
\lstset{style=mystyle}
\section{Can algorithms, as implemented in CLRS-30, have an equilibrium?}\label{app:equilibriums}

In this appendix, we have copied over some algorithms implementations from
CLRS-30\footnote{\url{https://github.com/google-deepmind/clrs/tree/master/clrs/_src/}}.
Additionally, we have annotated how and when they follow the $\ctexttt{while}\
b\ \ctexttt{do}\ c$ construct. Algorithms, that are \emph{not} necessarily
solved via this construct (e.g.\@ strongly connected components) were also
included, so as to showcase if this would break.

\begin{lstlisting}[caption={BFS algorithm. Clearly implemented as $\ctexttt{while}\ b\ \ctexttt{do}\ c$.}, label={app:lst:bfs}, language=Python]
def bfs(A: _Array, s: int) -> _Out:
  chex.assert_rank(A, 2)
  probes = probing.initialize(specs.SPECS['bfs'])
  A_pos = np.arange(A.shape[0])
  probing.push(
      probes,
      specs.Stage.INPUT,
      next_probe={
          'pos': np.copy(A_pos) * 1.0 / A.shape[0],
          's': probing.mask_one(s, A.shape[0]),
          'A': np.copy(A),
          'adj': probing.graph(np.copy(A))
      })
  reach = np.zeros(A.shape[0])
  pi = np.arange(A.shape[0])
  reach[s] = 1
  # Initialisation code ends here

  # implemented as do-while, but do-whiles are essentially 
  # c; while b do c
  while True:

    prev_reach = np.copy(reach)
    probing.push(
        probes,
        specs.Stage.HINT,
        next_probe={
            'reach_h': np.copy(prev_reach),
            'pi_h': np.copy(pi)
        })
    # command c: update reachability.
    for i in range(A.shape[0]):
      for j in range(A.shape[0]):
        if A[i, j] > 0 and prev_reach[i] == 1:
          if pi[j] == j and j != s:
            pi[j] = i
          reach[j] = 1

    if np.all(reach == prev_reach): # boolean condition b: has reachability vector changed
      break

  probing.push(probes, specs.Stage.OUTPUT, next_probe={'pi': np.copy(pi)})
  probing.finalize(probes)
  return pi, probes
\end{lstlisting}

\begin{lstlisting}[caption={Floyd-Warshall algorithm and its sampler (above). Can
also be viewed as $\ctexttt{while}\ b\ \ctexttt{do}\ c$, \emph{in CLRS-30}.},
label={app:lst:fw}, language=Python]
# The sampler
class FloydWarshallSampler(Sampler):
  """Sampler for all-pairs shortest paths."""

  def _sample_data(
      self,
      length: int,
      p: Tuple[float, ...] = (0.5,),
      low: float = 0., # never changed in the data generation
      high: float = 1., # never changed in the data generation
  ):
    graph = self._random_er_graph( # samples random ER graph with weights in [low;high)
        nb_nodes=length,
        p=self._rng.choice(p),
        directed=False,
        acyclic=False,
        weighted=True,
        low=low,
        high=high)
    return [graph]

# The implementation
def floyd_warshall(A: _Array) -> _Out:
  """Floyd-Warshall's all-pairs shortest paths (Floyd, 1962)."""

  chex.assert_rank(A, 2)
  probes = probing.initialize(specs.SPECS['floyd_warshall'])

  A_pos = np.arange(A.shape[0])

  probing.push(
      probes,
      specs.Stage.INPUT,
      next_probe={
          'pos': np.copy(A_pos) / A.shape[0],
          'A': np.copy(A),
          'adj': probing.graph(np.copy(A))
      })

  D = np.copy(A)
  Pi = np.zeros((A.shape[0], A.shape[0]))
  msk = probing.graph(np.copy(A))

  for i in range(A.shape[0]):
    for j in range(A.shape[0]):
      Pi[i, j] = i
  
  # Initialisation code ends here

  # for loops are while loops
  # ``for k in range(N)'' is equivalent to ``k:=0; while (i<N) do (c; k:=k+1)''
  # NOTE #1 The NN, however, has to learn to increase the k, solely
  # based on the `pos` input feature; having a `pred' feature,
  # (as in string algorithms) might have been more appropriate

  # NOTE #2 *Since the sampler (above) only samples positive edge weights*
  # no negative-weight cycles can exist. Hence any reruns of the inner two
  # for loops, after k iteraitons have passed will not change matrix D.

  # boolean condition b: k iterations have passed
  # (a necessary, *but not sufficient* condition is that D remains unchanged)
  for k in range(A.shape[0]): 
    prev_D = np.copy(D)
    prev_msk = np.copy(msk)

    probing.push(
        probes,
        specs.Stage.HINT,
        next_probe={
            'Pi_h': np.copy(Pi),
            'D': np.copy(prev_D),
            'msk': np.copy(prev_msk),
            'k': probing.mask_one(k, A.shape[0])
        })

    # command c: update D for intermediate vertex k
    for i in range(A.shape[0]):
      for j in range(A.shape[0]):
        if prev_msk[i, k] > 0 and prev_msk[k, j] > 0:
          if msk[i, j] == 0 or prev_D[i, k] + prev_D[k, j] < D[i, j]:
            D[i, j] = prev_D[i, k] + prev_D[k, j]
            Pi[i, j] = Pi[k, j]
          else:
            D[i, j] = prev_D[i, j]
          msk[i, j] = 1

  probing.push(probes, specs.Stage.OUTPUT, next_probe={'Pi': np.copy(Pi)})
  probing.finalize(probes)

  return Pi, probes
\end{lstlisting}

\begin{lstlisting}[caption={Kosaraju's strongly connected components \citep{sharir1981strong} algorithm. It is composed of \emph{four} (two nested ones, sequenced one after the other) $\ctexttt{while}\ b\ \ctexttt{do}\ c$ constructs.}, label={app:lst:scc}, language=Python]
def strongly_connected_components(A: _Array) -> _Out:
  """Kosaraju's strongly-connected components (Aho et al., 1974)."""

  chex.assert_rank(A, 2)
  probes = probing.initialize(
      specs.SPECS['strongly_connected_components'])

  A_pos = np.arange(A.shape[0])

  probing.push(
      probes,
      specs.Stage.INPUT,
      next_probe={
        # < omitted for brevity >
      })

  scc_id = np.arange(A.shape[0])
  color = np.zeros(A.shape[0], dtype=np.int32)
  d = np.zeros(A.shape[0])
  f = np.zeros(A.shape[0])
  s_prev = np.arange(A.shape[0])
  time = 0
  A_t = np.transpose(A)

  # Initialisation code ends here

  # boolean condition b: there are unvisited (color[s]=0) vertices
  for s in range(A.shape[0]):
    if color[s] == 0:
      s_last = s
      u = s
      v = s
      probing.push(
          probes,
          specs.Stage.HINT,
          next_probe={
            # < omitted for brevity >
          })
      # command c: grey them (color=1) and recursively visit descendants
      
      # NOTE command c is another while b' do c' with a stack
      while True: # b': stack is not empty 
        if color[u] == 0 or d[u] == 0.0:
          time += 0.01
          d[u] = time
          color[u] = 1
          probing.push(
              probes,
              specs.Stage.HINT,
              next_probe={
                # < omitted for brevity >
              })
        for v in range(A.shape[0]): # c': add lowest id unvisited descendant of top-stack node to the top of the stack
          if A[u, v] != 0:
            if color[v] == 0:
              color[v] = 1
              s_prev[v] = s_last
              s_last = v
              probing.push(
                  probes,
                  specs.Stage.HINT,
                  next_probe={
                    # < omitted for brevity >
                  })
              break

        if s_last == u: # no descending
          color[u] = 2
          time += 0.01
          f[u] = time

          probing.push(
              probes,
              specs.Stage.HINT,
              next_probe={
                # < omitted for brevity >
              })

          if s_prev[u] == u: # and we are on top of the recursion
            # although imaginary, in the implementation here,
            # if a stack was used, it'd be empty in this if
            # statement
            assert s_prev[s_last] == s_last
            break
          pr = s_prev[s_last]
          s_prev[s_last] = s_last
          s_last = pr

        u = s_last

  color = np.zeros(A.shape[0], dtype=np.int32)
  s_prev = np.arange(A.shape[0])

  # boolean condition b'': there are unvisited (color[s]=0) vertices
  # (order of visiting depends on finishing time;
  #  see Introduction to algorithms, 4th edition, Chapter 20)
  for s in np.argsort(-f):
    if color[s] == 0:
      s_last = s
      u = s
      v = s
      probing.push(
          probes,
          specs.Stage.HINT,
          next_probe={
            # < omitted for brevity >
          })
      # NOTE command c is another while b''' do c''' with a stack
      while True: # b''': stack is not empty 
        scc_id[u] = s
        if color[u] == 0 or d[u] == 0.0:
          time += 0.01
          d[u] = time
          color[u] = 1
          probing.push(
              probes,
              specs.Stage.HINT,
              next_probe={
                # < omitted for brevity >
              })
        for v in range(A.shape[0]): # c''': add lowest id unvisited descendant of top-stack node to the top of the stack
          if A_t[u, v] != 0:
            if color[v] == 0:
              color[v] = 1
              s_prev[v] = s_last
              s_last = v
              probing.push(
                  probes,
                  specs.Stage.HINT,
                  next_probe={
                    # < omitted for brevity >
                  })
              break

        if s_last == u:
          color[u] = 2
          time += 0.01
          f[u] = time

          probing.push(
              probes,
              specs.Stage.HINT,
              next_probe={
                # < omitted for brevity >
              })

          if s_prev[u] == u: # same as before
            assert s_prev[s_last] == s_last
            break
          pr = s_prev[s_last]
          s_prev[s_last] = s_last
          s_last = pr

        u = s_last

  probing.push(
      probes,
      specs.Stage.OUTPUT,
      next_probe={'scc_id': np.copy(scc_id)},
  )
  probing.finalize(probes)

  return scc_id, probes
\end{lstlisting}

\section{Differences to \citet{ibarz2022generalist}}\label{app:differencesCLRS}

Our differences are mostly required by software engineering rather than
research, hence they live here. Differences are: 
\begin{itemize}
    \item Different DL framework (Pytorch \citep{paszke2019pytorch})
    \item \citet{ibarz2022generalist} use an extra nonlinearity after
        the GNN step. We found this to be not necessary (there are
        plenty of nonlinearities at the message function) for the baseline
        and to be making the training of DEARs less stable so we removed it.
    \item Sorting-based algorithms use a Sinkhorn operator to force the output to
        be a permutation. However, this gave very negative logits for the predictions
        at initialisation, leading to runs starting from a very high loss and
        converging to poorer minima. We fixed this by adding an off-centred
        leaky ReLU activation with the kink point at (-6, -6) right after the
        Sinkhorn operator. After conversion of logits to outputs via softmax,
        our change is mathematically equivalent to saying that the probability for each
        other node to be predecessor should not drop below $10^{-6}$.
\end{itemize}

\section{Picking least fixed point}\label{app:lfp}

For a given batch fixed point finding continues until all instances
in the batch converge and \emph{at each step} the solver is stepped \emph{on
all} instances.  For a given instance, when two $\rmH^{(t)}$ and $\rmH^{(t')}$
are under the threshold $\delta$, for some $t\leq t'$, the \texttt{torchdeq}
library prefers the state that has the lower distance to next state.
Consequently, out the returned fixed points only one is guaranteed to be least
-- the one that require the most steps. This not only misaligns with domain
theory, but also had the practical effect that the neural models require more
iterations to converge the more we train them. Thus, we changed the library to
choose the first $\rmH{(t)}$ that passes the fixed point criteriaq.

\section{Alignment algorithm}\label{app:alignDP}

Assume we have computed pairwise distance matrices between the states and those
are stored in a $T\times T_\mathcal{G}$ distance matrix $\rmD$ with elements
$d_{i,j}$. Ignoring the required alignment of the last states, we focus on
aligning the rest of the states. This is done via standard dynamic programming
algorithm. The dynamic programming state we define is as follows: $dp_{i,j}$ is
the most optimal alignment for the first $i$ DEAR states and first $j$ NAR
states, with $dp_{0,j}=0$ (having leftover NAR states mean we skipped some, but
we do not want to penalise that) and $dp_{i,0}=\infty$ (we want to align all
DEAR states). We consider two recursive formulas, first one we use, the other
we use when $T\leq T_\mathcal{G}$:
\begin{itemize}
    \item when the $T>T_\mathcal{G}$, there are extra states. To avoid
        infinities we will allow for two DEAR states to align to a same state:
        \begin{align}
            dp_{i,j}= \min
                \begin{cases}
                dp_{i-1,j}+d_{i,j} & \text{aligning DEAR state $i$ and NAR state $j$, but allowing for}\\
                                   & \text{previous states to align to it as well}\\
                    dp_{i,j-1} & \text{skipping alignment with state $j$}
                \end{cases}
        \end{align}
    \item when the $T\leq T_\mathcal{G}$ we require that each DEAR state aligns to an unique NAR state:
        \begin{align}
            dp_{i,j}= \min
                \begin{cases}
                    dp_{i-1,\textcolor{Purple}{j-1}}+d_{i,j} & \text{aligning DEAR state $i$ and NAR state $j$}\\
                    dp_{i,j-1} & \text{skipping alignment with state $j$}
                \end{cases}
        \end{align}
        We have highlighted the difference to the above in \textcolor{Purple}{purple}.
\end{itemize}
The optimal alignment for the whole two sequence is stored in
$dp_{T,T_\mathcal{G}}$. As both $\rmH^{(0)}$ and $\rmH^{(0)}_\mathcal{G}$ are
concatenation of 0 vectors (due to how we initialise the latent state), their
distance is always 0 and they will always align as required. To enforce
alignment of the last state, we take the optimal value for the subsequences
without last states $dp_{T-1,T_\mathcal{G}-1}$ and \emph{always} (even when
subsampling, see below) add the distances between the last states to the loss
function.

As the above will always penalise longer DEQ trajectories, we divide $dp_{T-1,
T_\mathcal{G}-1}$ by $T-1$ before including it in the loss function. Lastly, to
allow for ``intermediate'' states (ones not necessarily matching a GNN state) to
exist, we subsample randomly, without replacement,
$T'=\max(\floor{\frac{T-1}{2}}, 1)$ DEAR states and apply the dynamic programming
algorithm with the subsampled sequence.



\clearpage

\section{Training loss: NAR vs DEAR}\label{app:NARvsDEARindividuals}

\begin{figure}[h]
    \centering
    \begin{subfigure}{0.4\linewidth}
        \includegraphics[width=\linewidth]{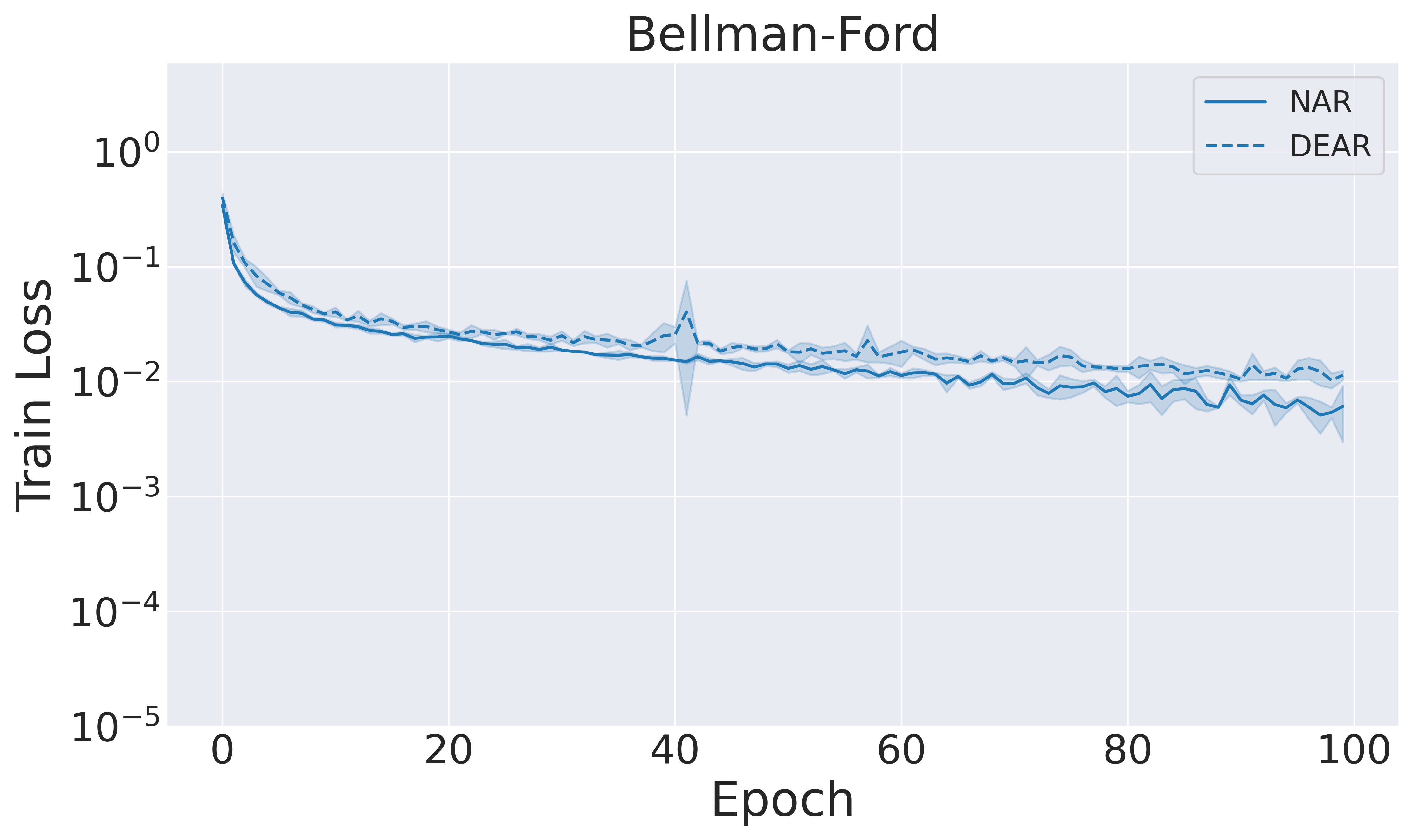}
    \end{subfigure}
    \begin{subfigure}{0.4\linewidth}
        \includegraphics[width=\linewidth]{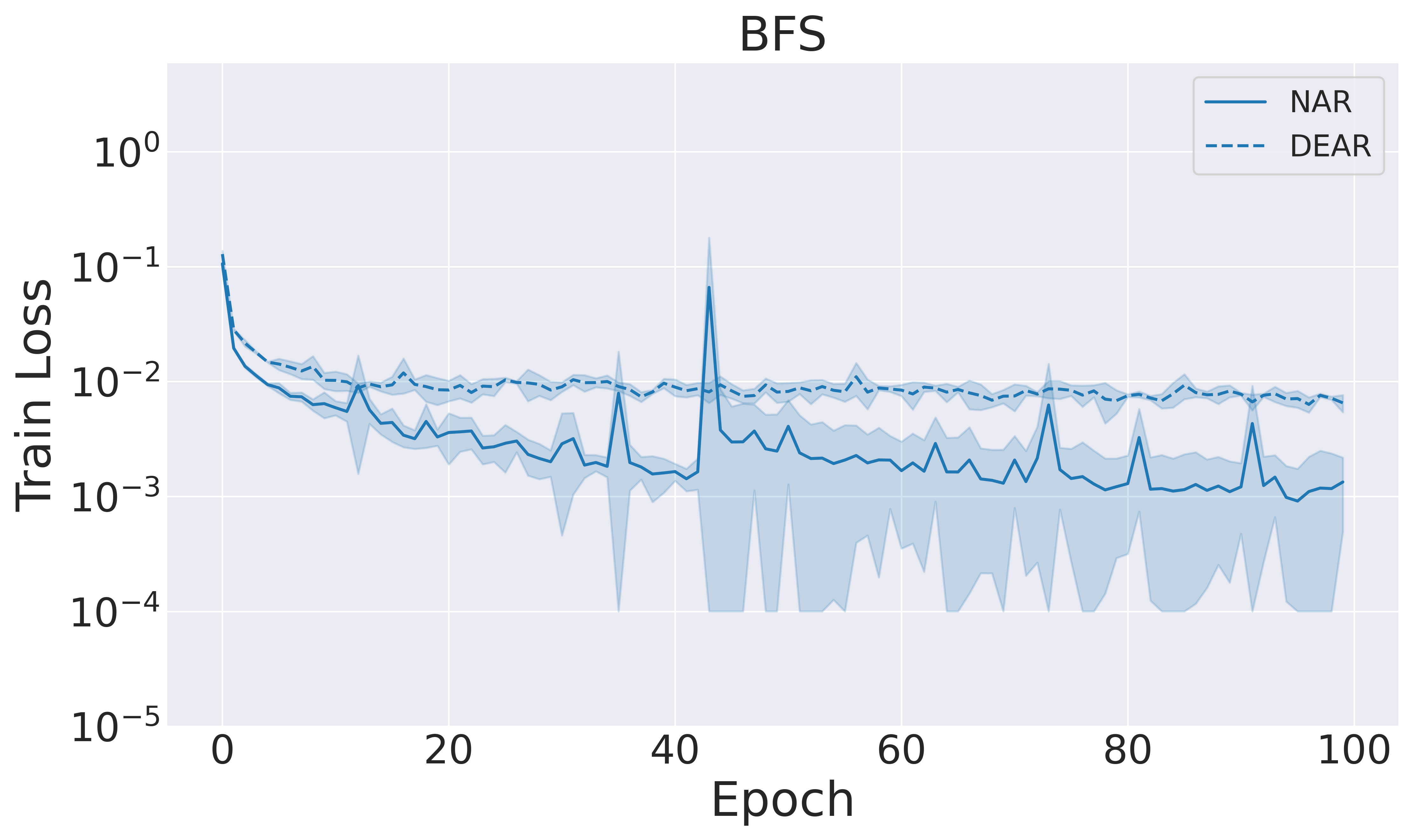}
    \end{subfigure}
    \begin{subfigure}{0.4\linewidth}
        \includegraphics[width=\linewidth]{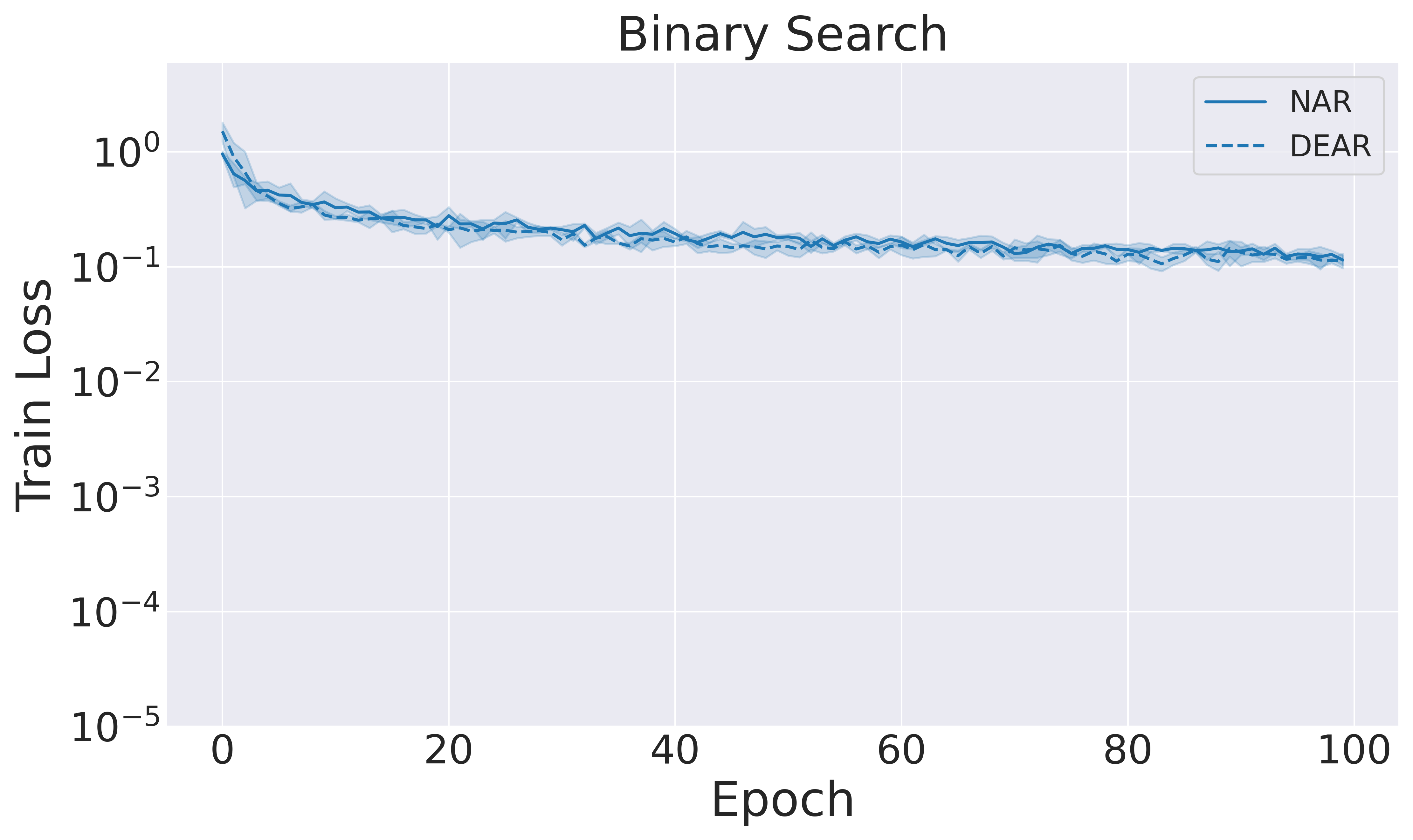}
    \end{subfigure}
    \begin{subfigure}{0.4\linewidth}
        \includegraphics[width=\linewidth]{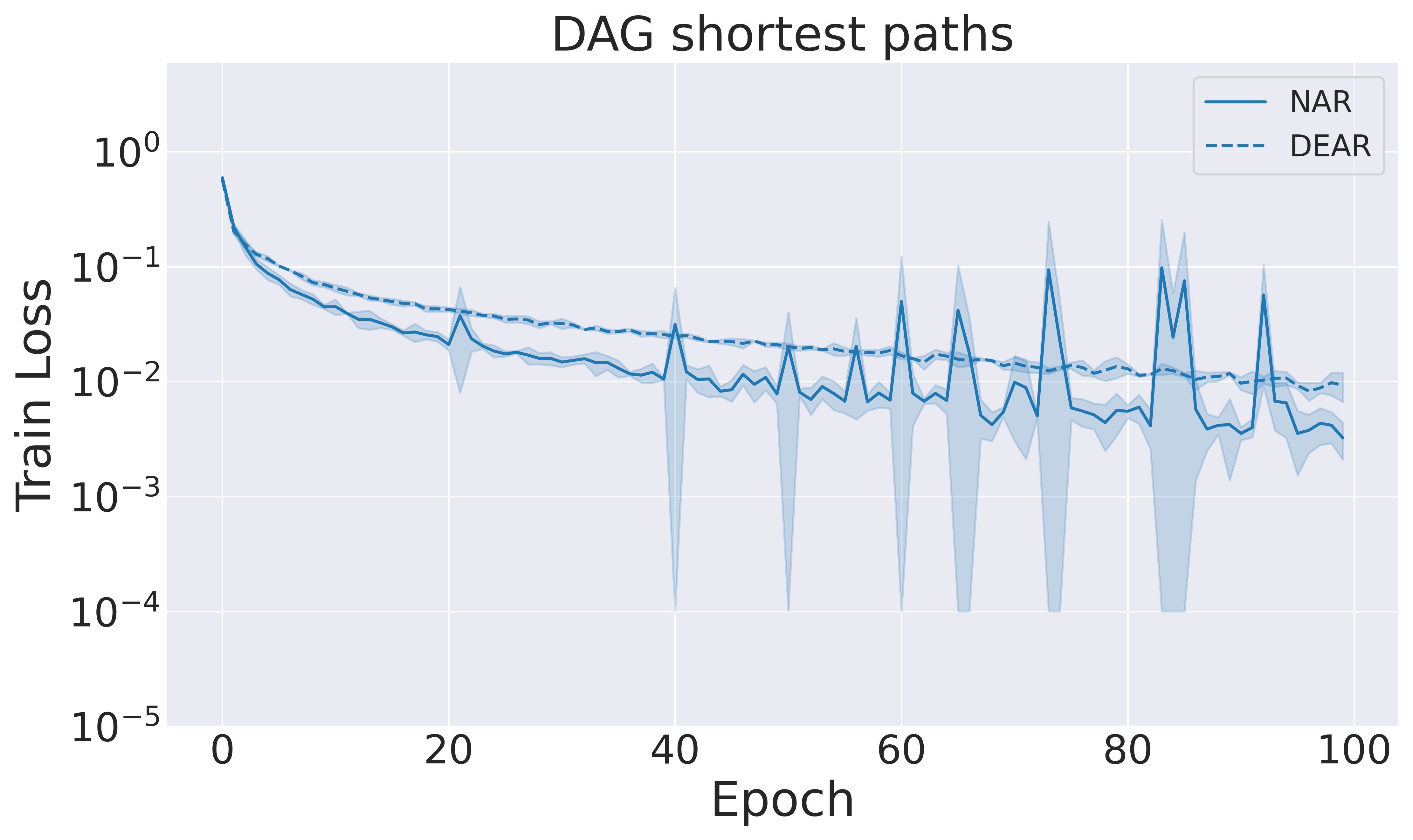}
    \end{subfigure}
    \begin{subfigure}{0.4\linewidth}
        \includegraphics[width=\linewidth]{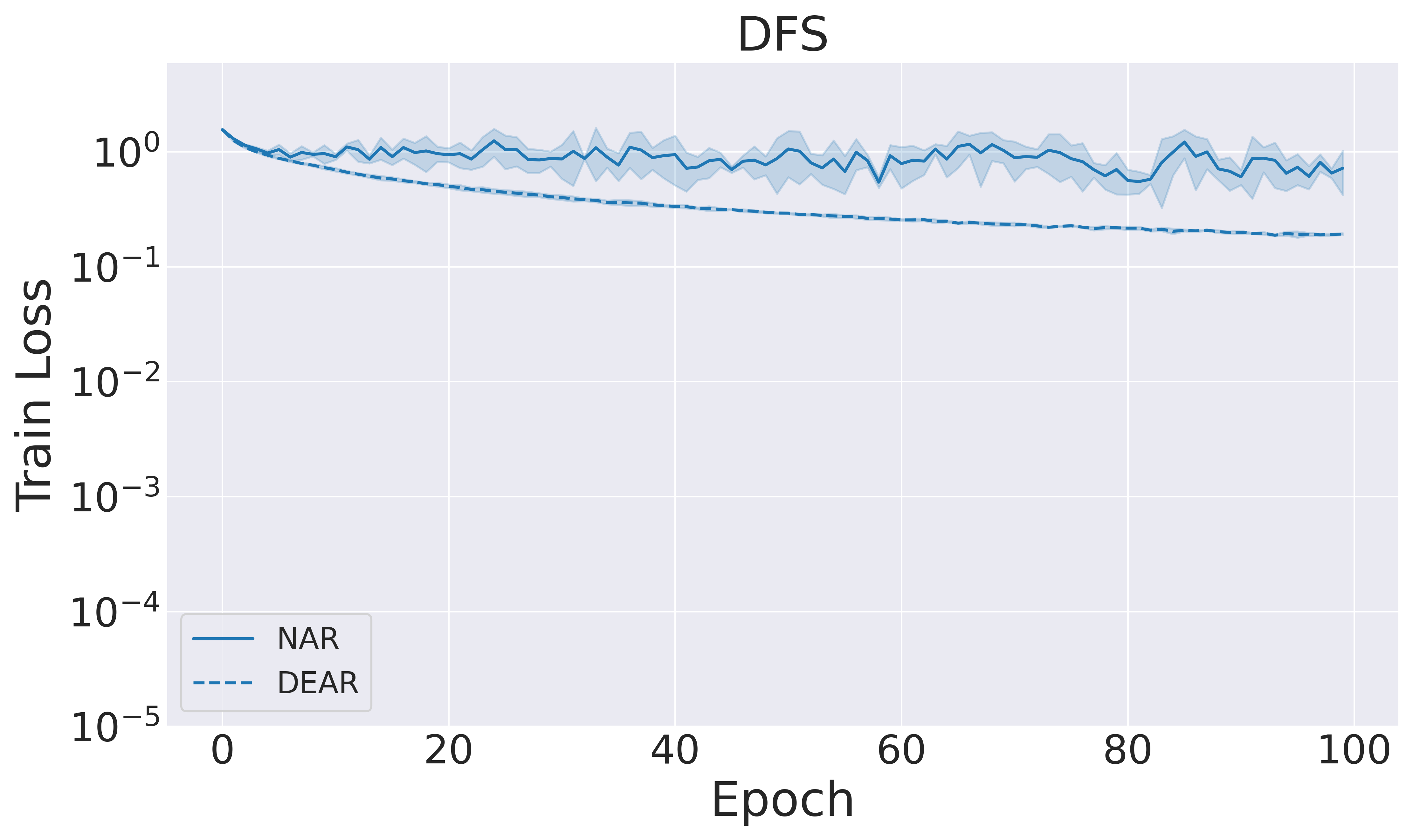}
    \end{subfigure}
    \begin{subfigure}{0.4\linewidth}
        \includegraphics[width=\linewidth]{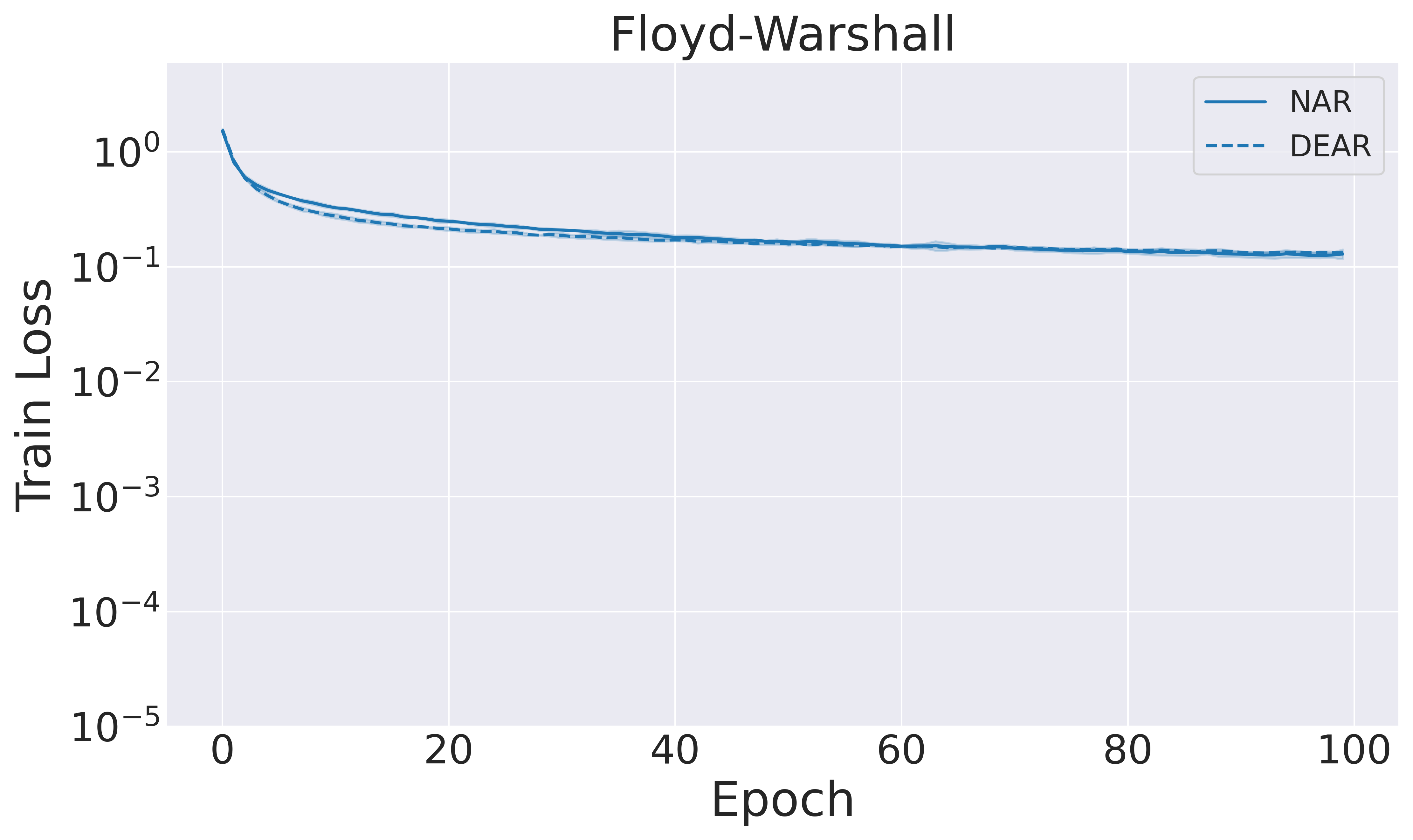}
    \end{subfigure}
    \begin{subfigure}{0.4\linewidth}
        \includegraphics[width=\linewidth]{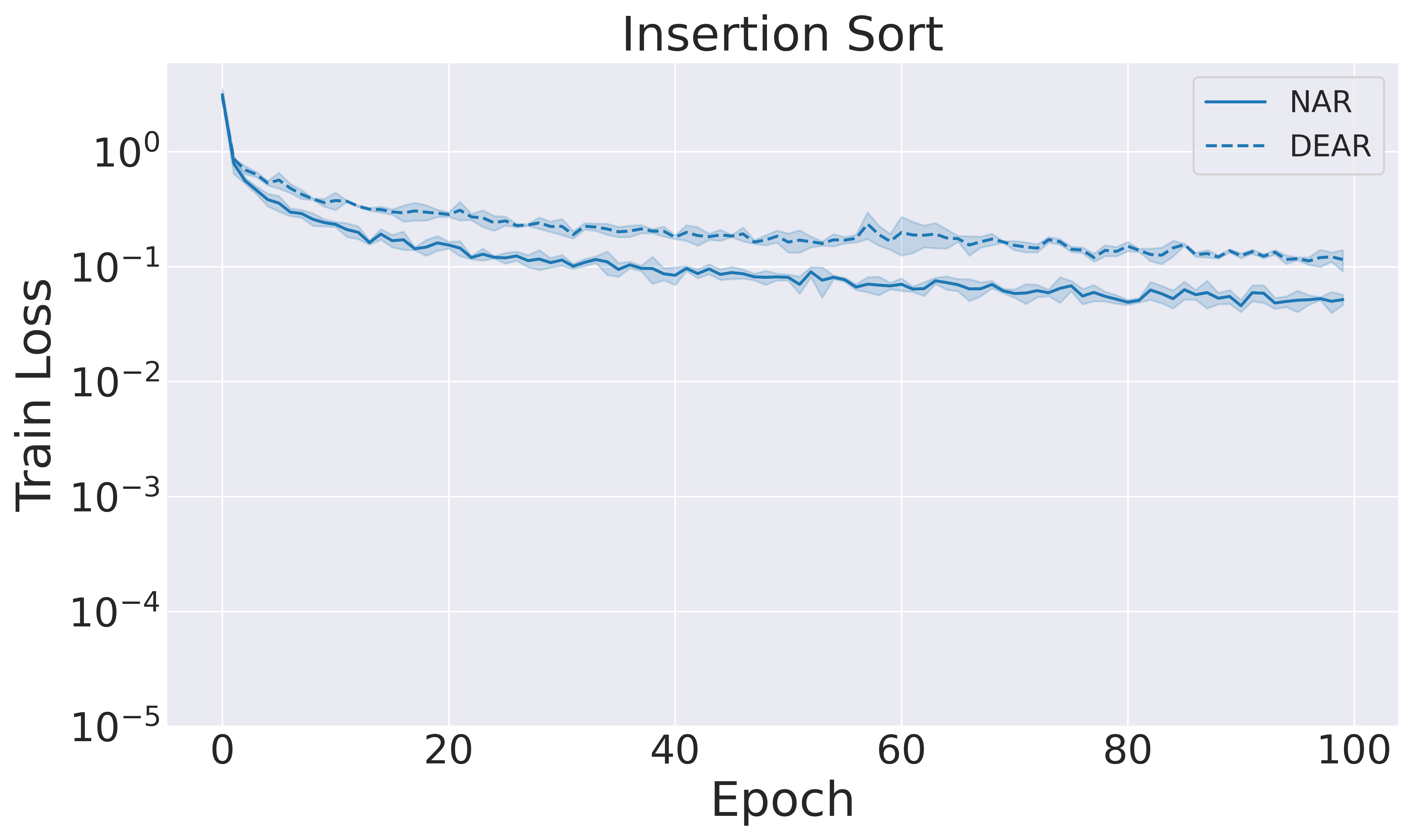}
    \end{subfigure}
    \begin{subfigure}{0.4\linewidth}
        \includegraphics[width=\linewidth]{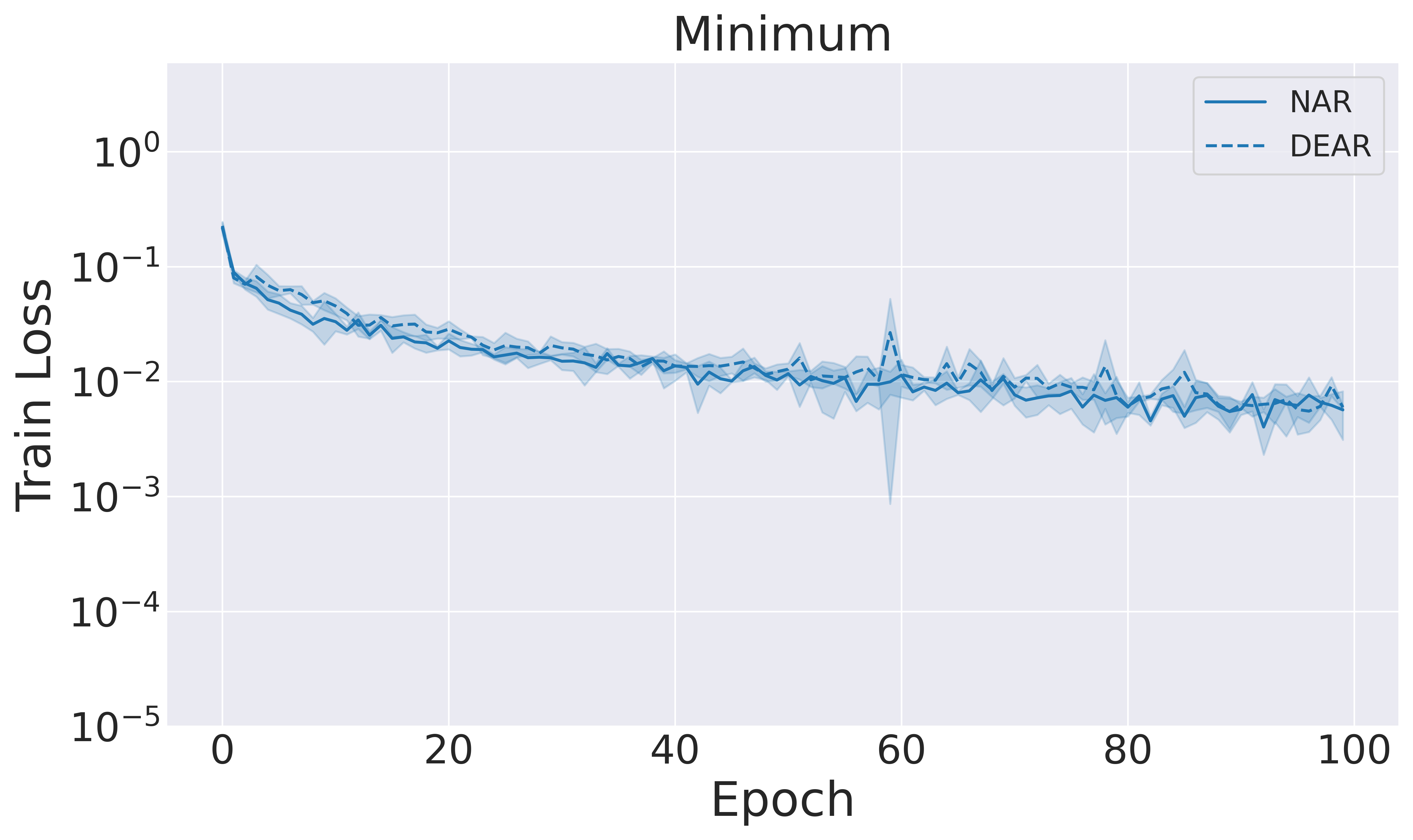}
    \end{subfigure}
    \begin{subfigure}{0.4\linewidth}
        \includegraphics[width=\linewidth]{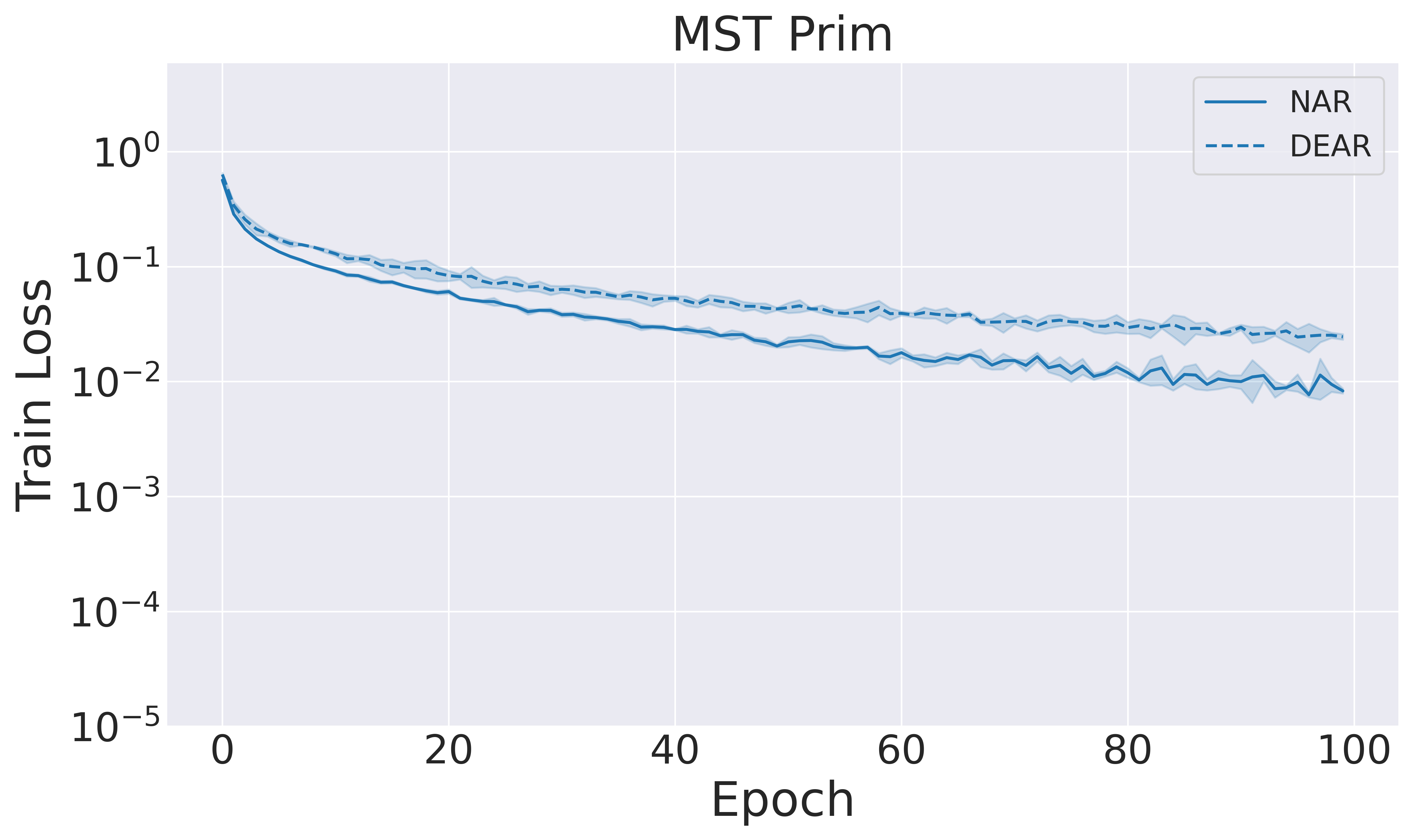}
    \end{subfigure}
    \begin{subfigure}{0.4\linewidth}
        \includegraphics[width=\linewidth]{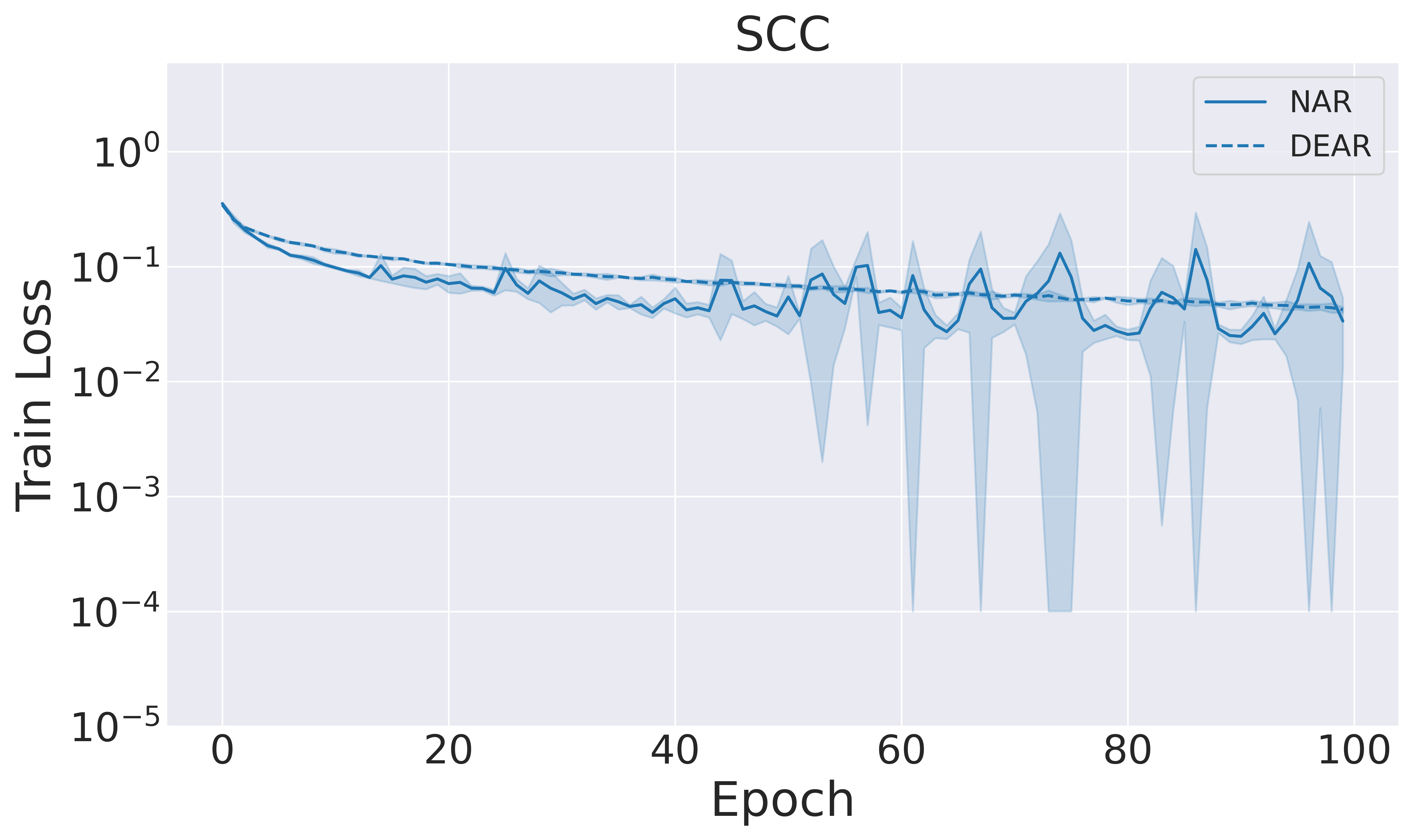}
    \end{subfigure}
    \caption{Side-by-side comparison of NAR vs DEAR. DEAR training loss is always within 1 order of magnitude of NAR. Note the log scale.}
\end{figure}
\clearpage

\section{Binary search anomalies}\label{app:BSano}

In the CLRS-30 implementation of binary search, we aim to find the place to
insert the target value \texttt{x} in a sorted array \texttt{A}. Thus we need
to point to the graph node that holds the smallest value in the array
\texttt{A}, which is greater than \texttt{x}. However, if \texttt{x>max (A)}, the
answer is a pointer to the last value of the array, which by the convention
used by CLRS-30 means we would be inserting \texttt{x} at the wrong place. In
other words, the answer to \texttt{A=[0.1, 0.2, 0.3] x=0.25} and
\texttt{A=[0.1, 0.2, 0.3] x=0.35} is the same -- insert \texttt{x} to the left
of 0.3. This contributed some noise, so we fixed the sampler to always give
\texttt{x} within \texttt{[0, max(A))}.  The other changes were to explicitly
use \texttt{graph \& pointer} instead of \texttt{node \& mask\_one} as the
location \& datatype of the pointer to the position in the array, also done by
\citet{engelmayer2023parallel}. Similarly to them, we also add an additional
supervision signal, but at the output level rather than the hint level --
teaching the models to predict which array elements are smaller than
\texttt{x} (binary \texttt{mask} type).

\begin{table}[t]
\centering
\caption{Fixing anomalies with CLRS-30's binary search further increases our overall score making our approach very competitive to Triplet-MPNN. Notation taken from \autoref{tab:OHDEAR}.}\label{chap6:tab:fixed_bs}
\footnotesize
\begin{tabular}{lcccccc}
\toprule
    Algorithm & \textbf{NAR$^{\blacklozenge}$} & \specialcell{\textbf{NAR$^{\blacklozenge}$}\\{(Triplet-MPNN)}} & \specialcell{\textbf{NAR$^{\diamondsuit}$}\\{(LT)}} & \specialcell{\textbf{DEAR}\\(ours)} \\
\midrule
\textbf{Search} (Parallel) & $95.67\% \pm 0.58$ & $93.33\%\pm0.58$ & $93.33\%\pm3.05$ & $85.67\% \pm 0.58$ \\
\midrule
{\textbf{New Overall}} & $71.52\%$ & $77.58\%$ & $70.97\%$ & $78.38\%$ \\
\bottomrule
\end{tabular}
\end{table}

We reran the new, parallel version of search, reporting results in
\autoref{chap6:tab:fixed_bs}. Our model still falls short of the baselines, but
the 26\% increase in accuracy is large enough to give a slight overall
advantage to DEAR over the Triplet-MPNN model. We do note, however, that the
task of searching is mostly numerical (comparison between floating point
numbers), resulting in DEAR overfitting a lot -- recall that train accuracy was
95\% even for the original (binary) search. We verified that if the training
data is increased $3\times-5\times$, test accuracy becomes comparable to other
models, regardless of which version is used.

\clearpage
\section{Training loss: DEAR vs DEAR w/ CGP}\label{app:DEARvsCGPindividuals}

\begin{figure}[h]
    \centering
    \begin{subfigure}{0.4\linewidth}
        \includegraphics[width=\linewidth]{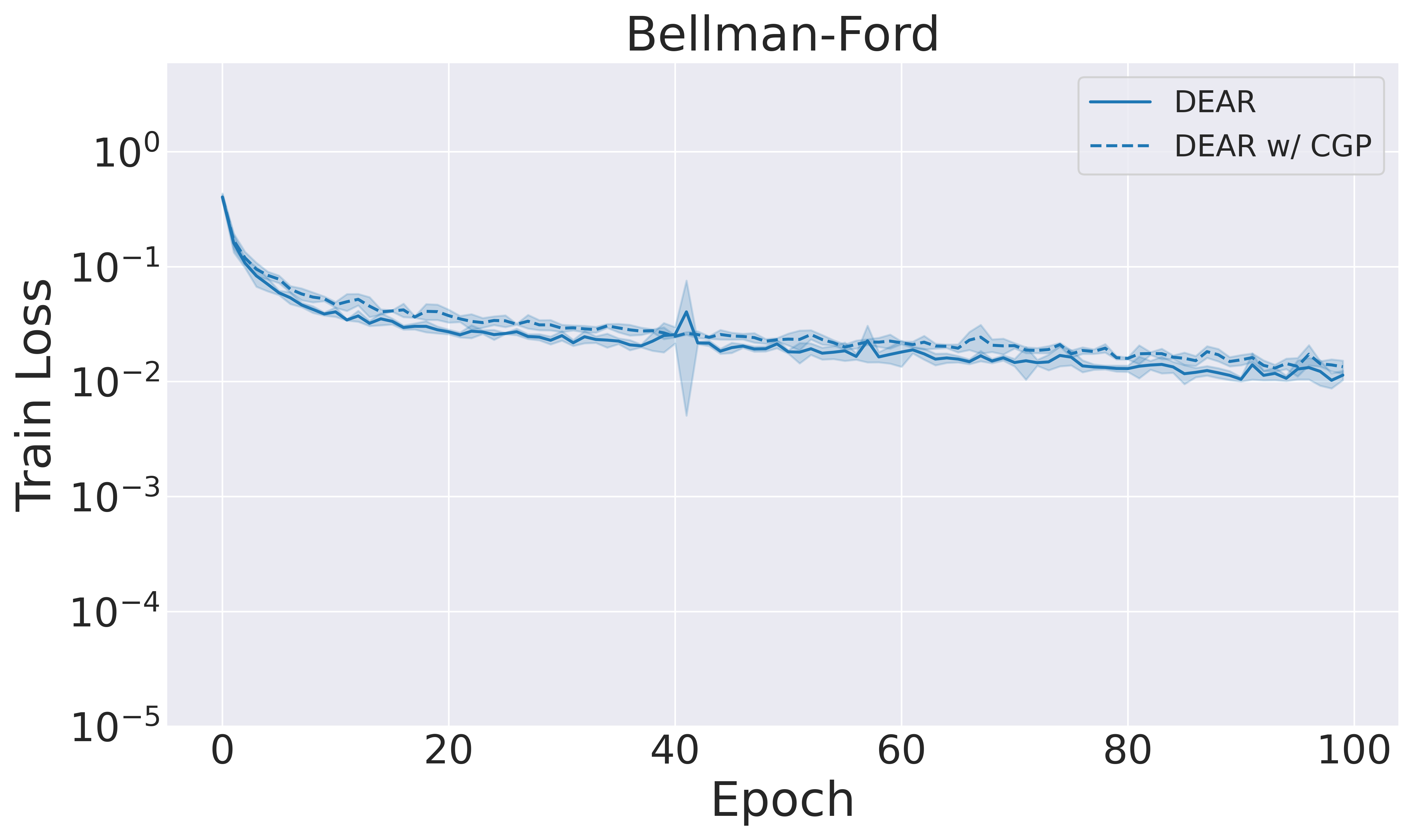}
    \end{subfigure}
    \begin{subfigure}{0.4\linewidth}
        \includegraphics[width=\linewidth]{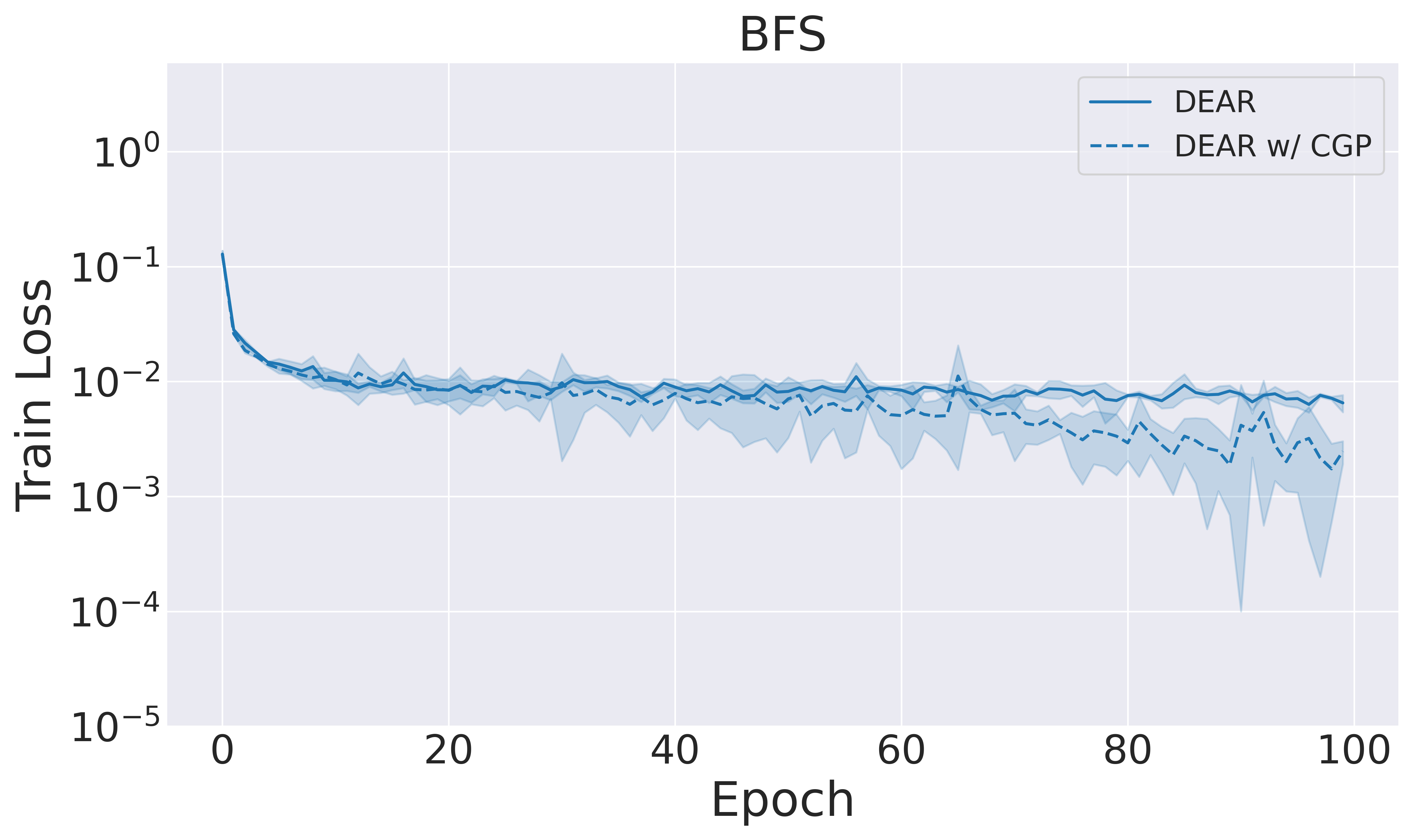}
    \end{subfigure}
    \begin{subfigure}{0.4\linewidth}
        \includegraphics[width=\linewidth]{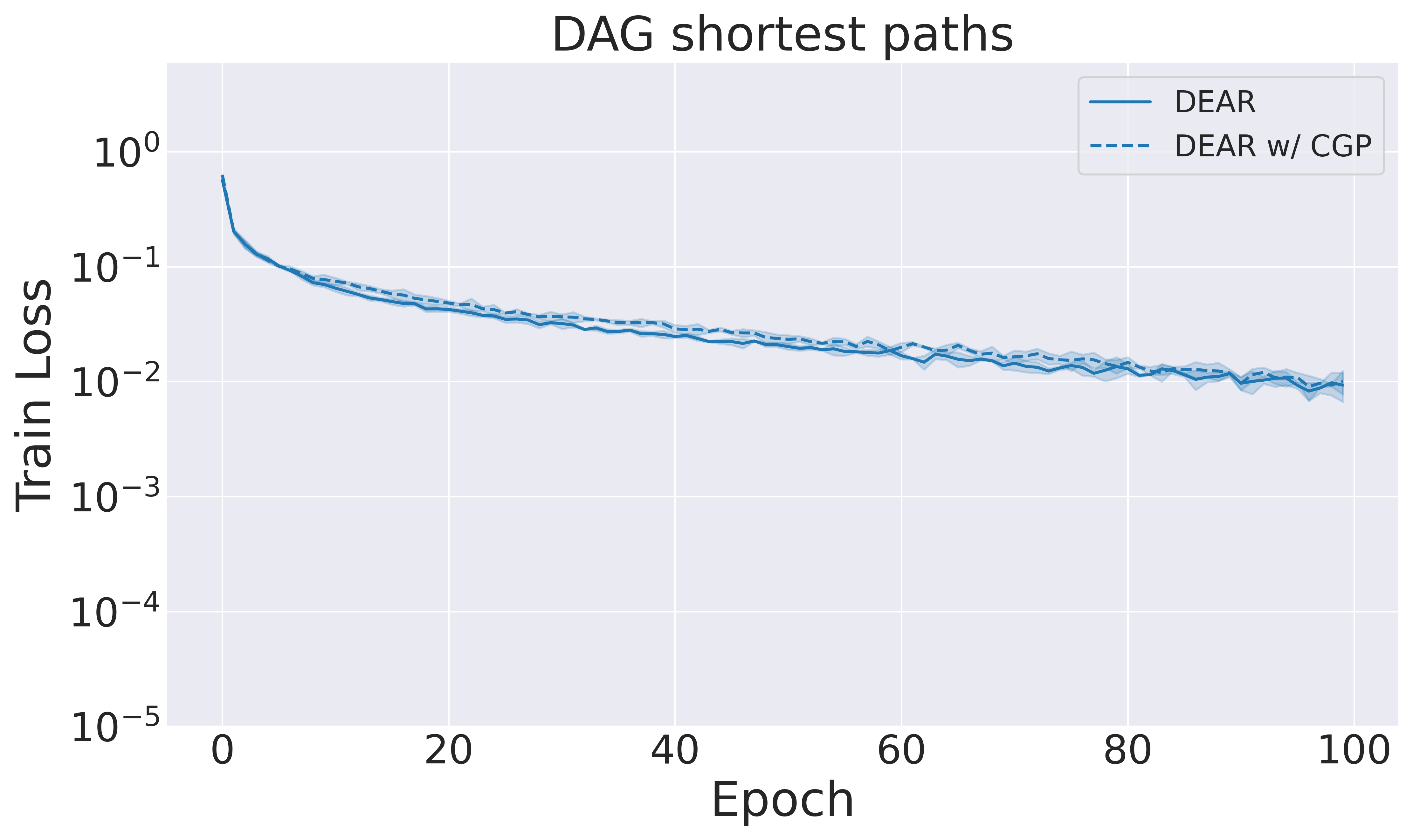}
    \end{subfigure}
    \begin{subfigure}{0.4\linewidth}
        \includegraphics[width=\linewidth]{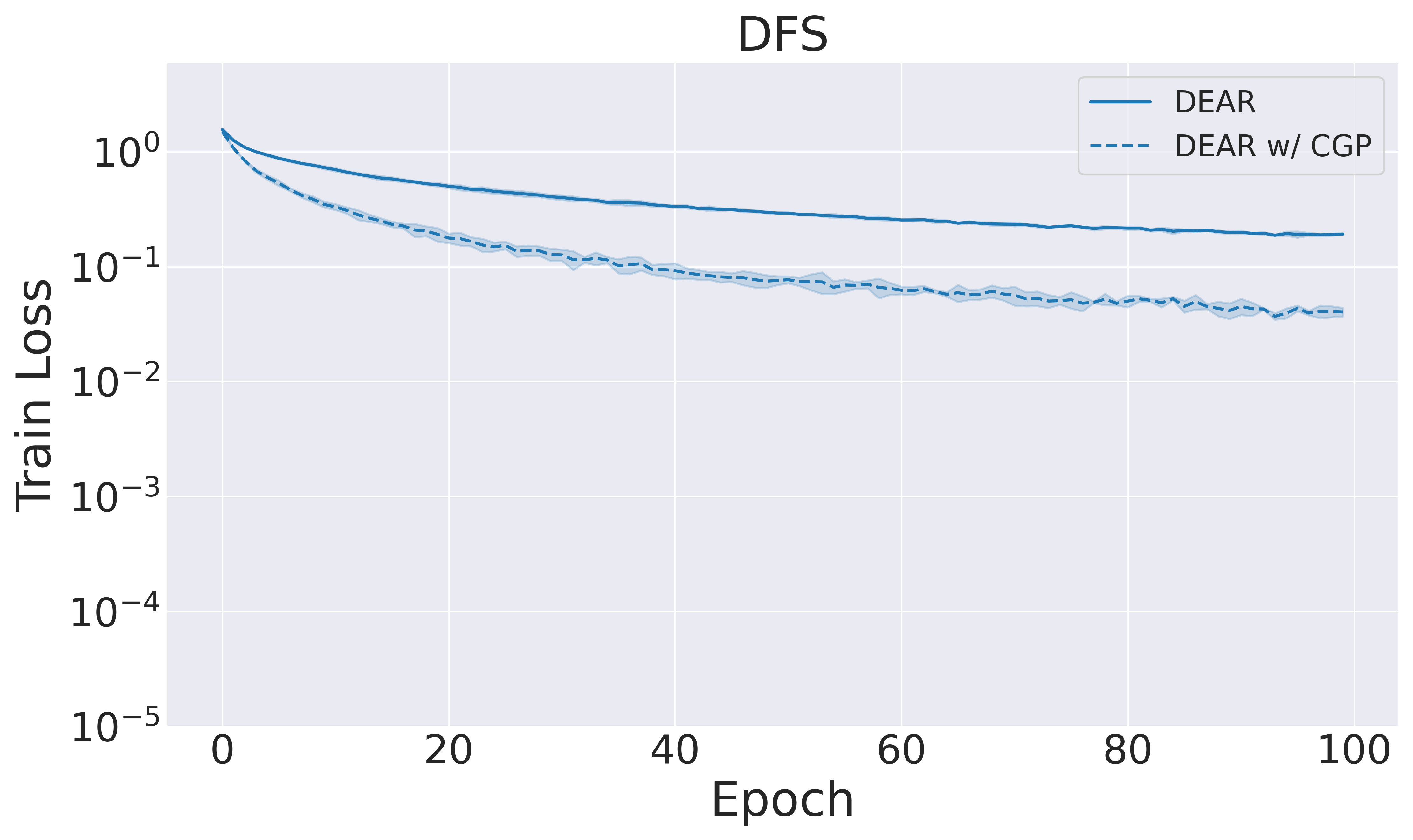}
    \end{subfigure}
    \begin{subfigure}{0.4\linewidth}
        \includegraphics[width=\linewidth]{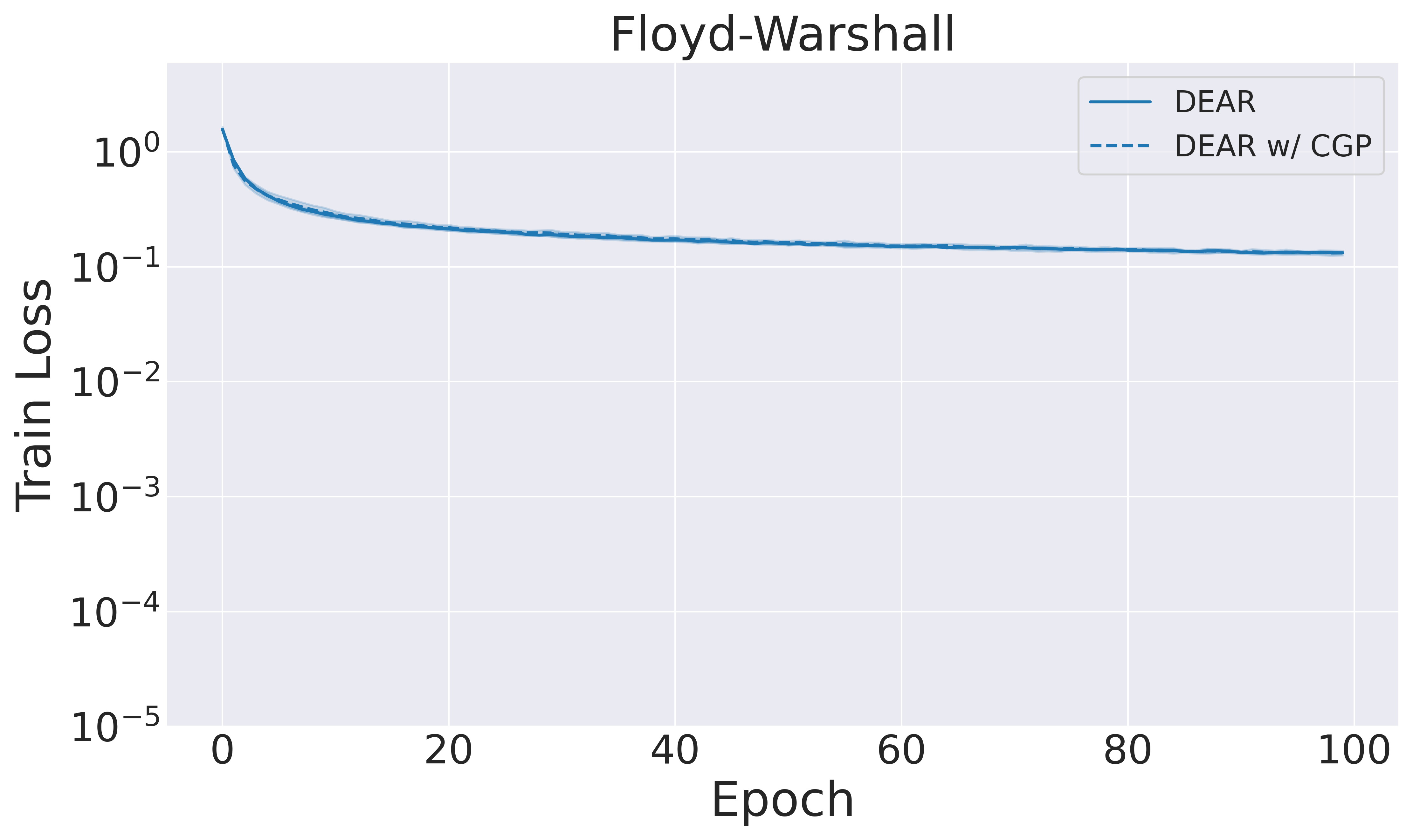}
    \end{subfigure}
    \begin{subfigure}{0.4\linewidth}
        \includegraphics[width=\linewidth]{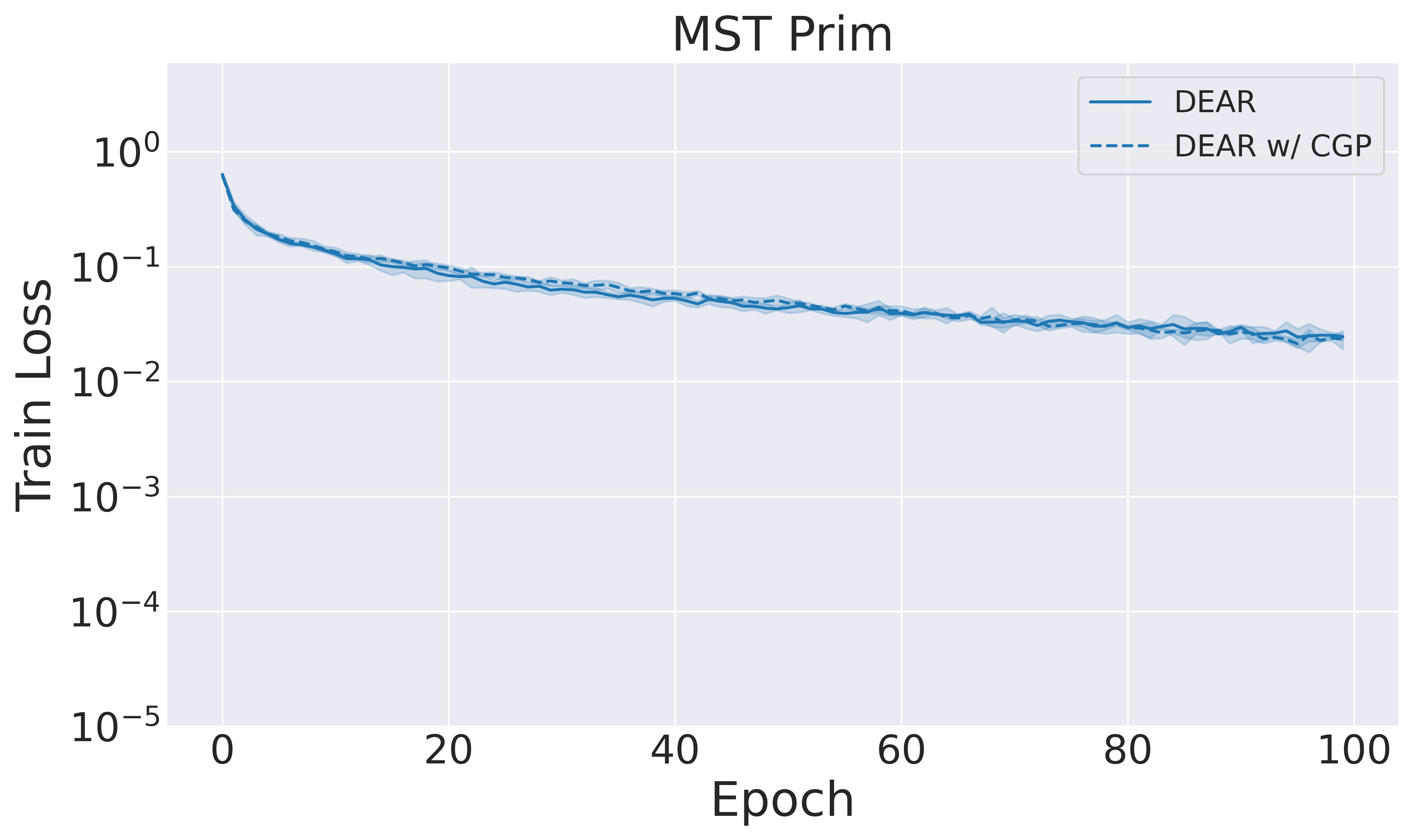}
    \end{subfigure}
    \begin{subfigure}{0.4\linewidth}
        \includegraphics[width=\linewidth]{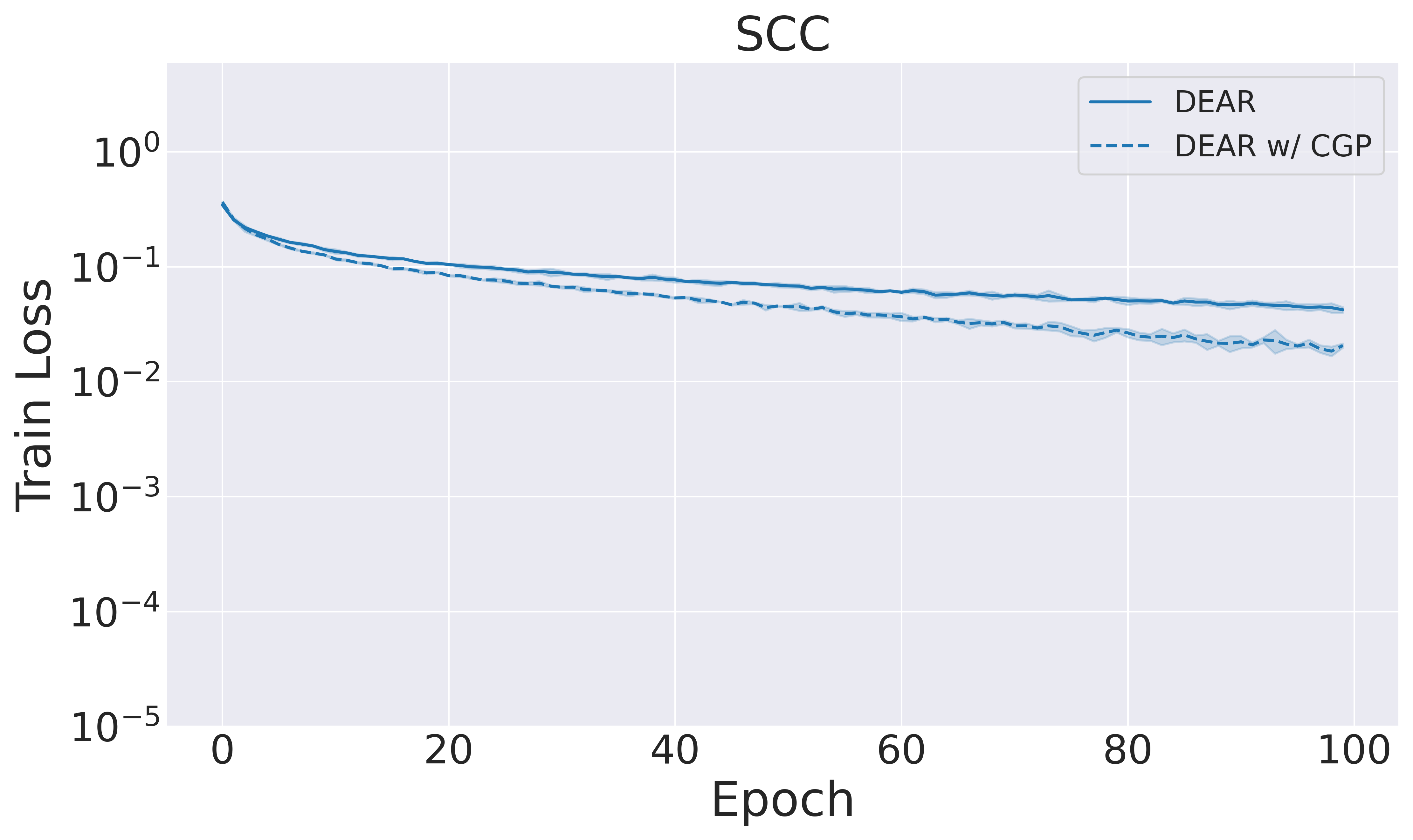}
    \end{subfigure}
    \caption{Effect of using Cayley Graph propagation on the train loss.}
\end{figure}

\clearpage
\section{Alignment gives closer convergence}\label{app:alignftw}

\begin{figure}[h]
    \centering%
    \begin{subfigure}{0.65\linewidth}
        \includegraphics[width=1\linewidth]{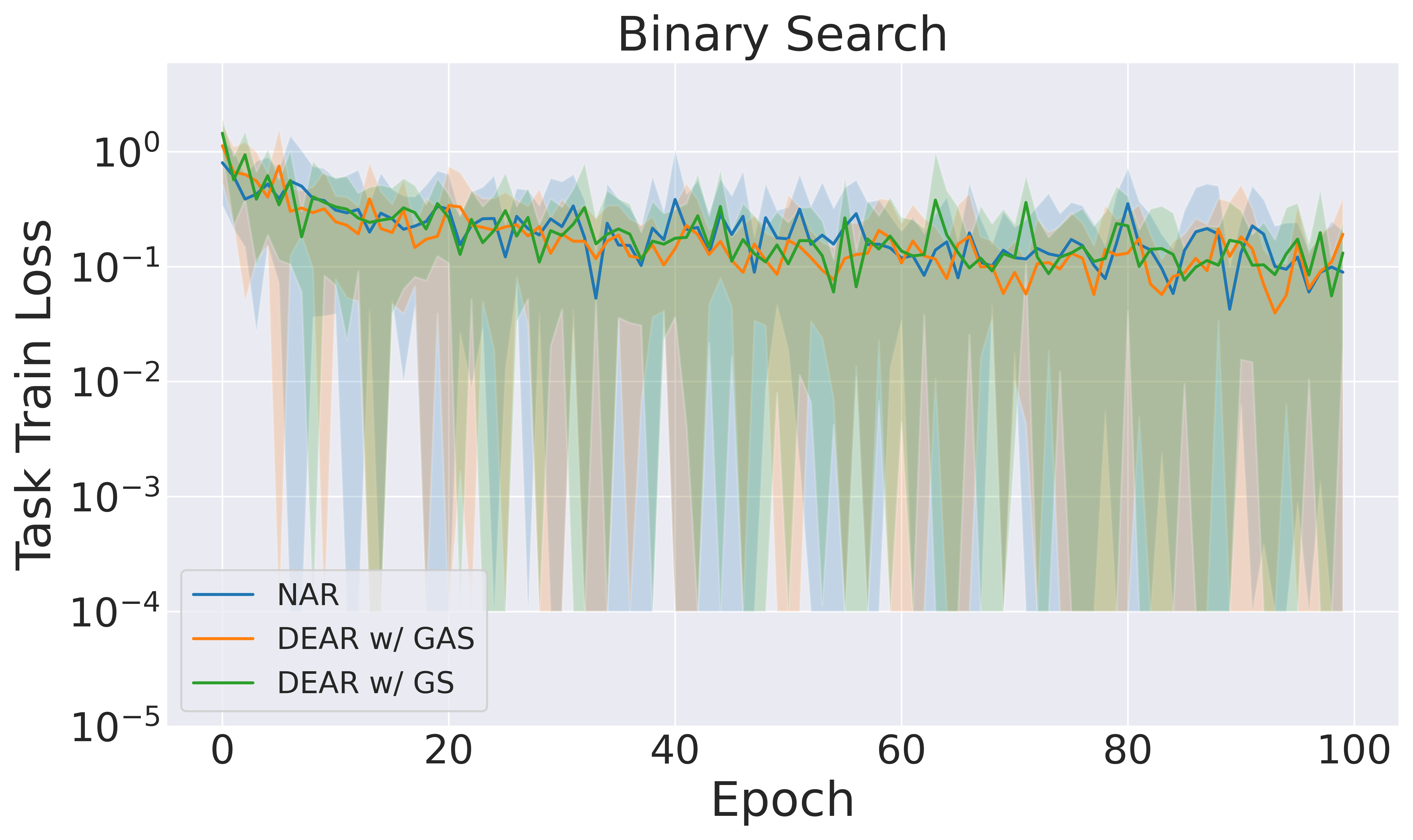}
    \end{subfigure}
    \begin{subfigure}{0.65\linewidth}
        \includegraphics[width=1\linewidth]{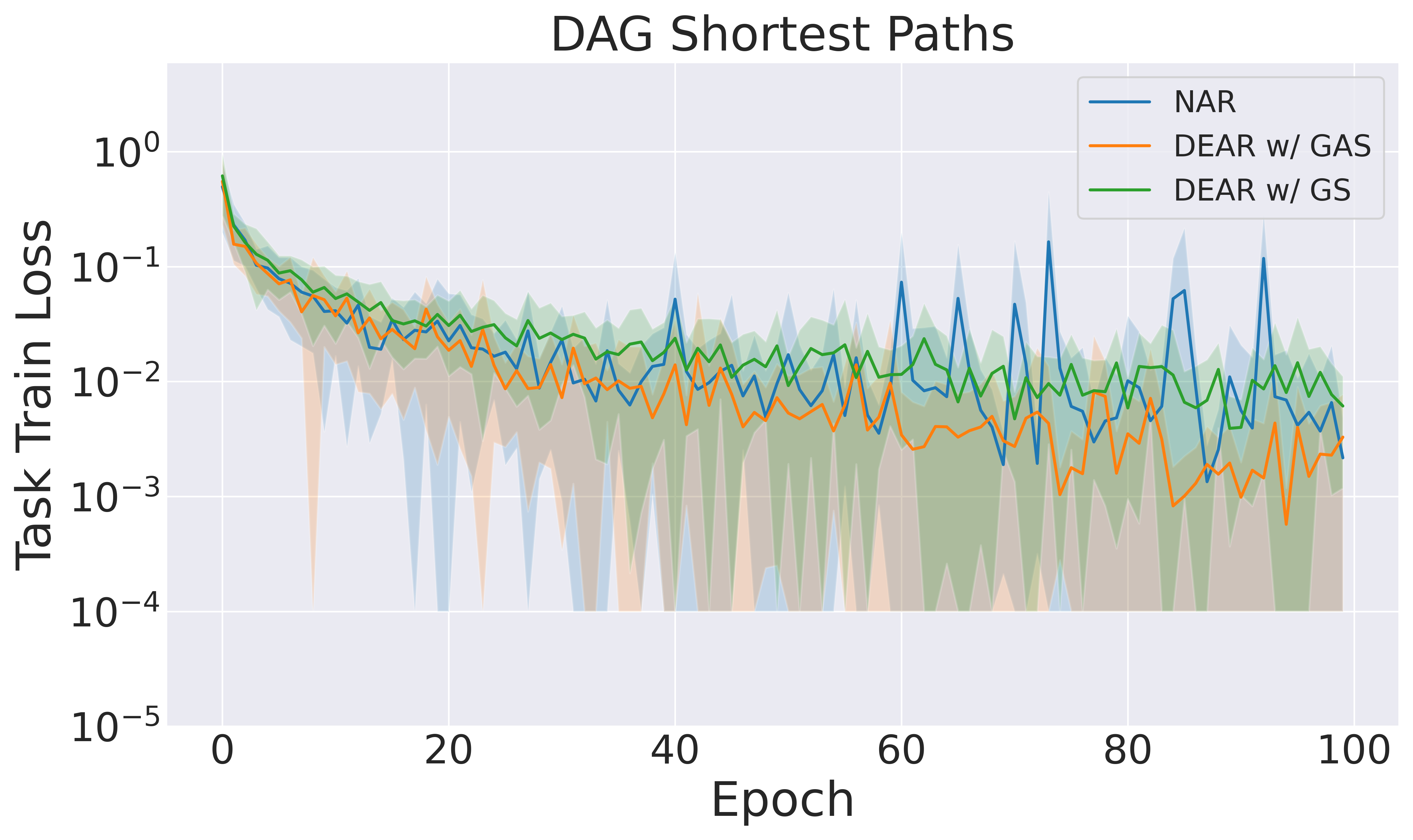}
    \end{subfigure}
    \begin{subfigure}{0.65\linewidth}
        \includegraphics[width=1\linewidth]{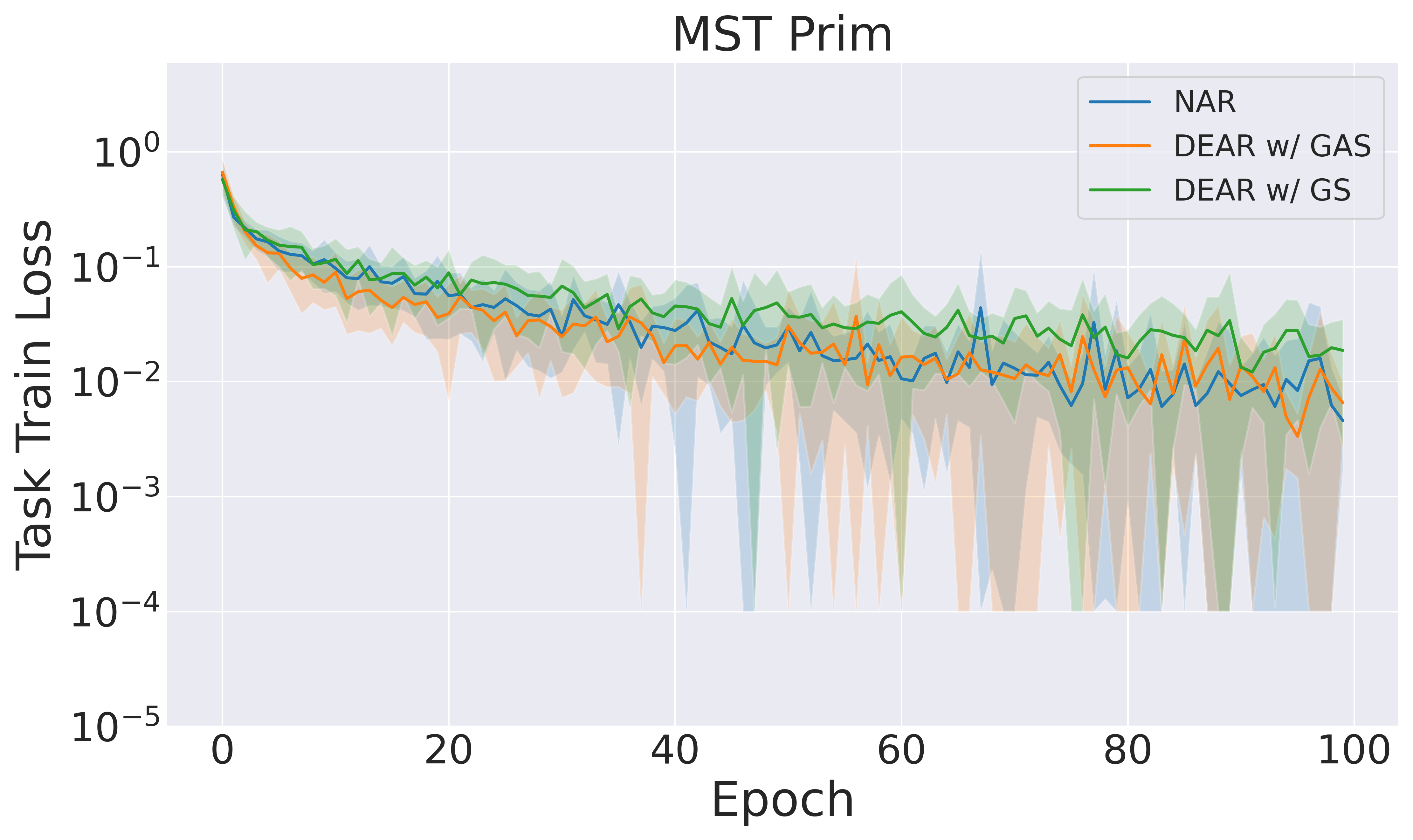}
    \end{subfigure}
    \caption{Alignment (orange) leads to lower task train loss compared to no aligment, but using stochasticity and GRANOLA (green).}\label{fig:amenoAlignment}
\end{figure}


\end{document}